\documentclass{article}
\usepackage[utf8]{inputenc}
\usepackage{natbib}%
\usepackage[figuresright]{rotating}
\usepackage{geometry}

\usepackage{url}
\usepackage{amsmath,amssymb,amsfonts}
\usepackage{empheq}
\usepackage[algoruled, lined, linesnumbered]{algorithm2e}
\usepackage{algcompatible}
\usepackage[utf8]{inputenc}
\usepackage{color}
\usepackage{multirow,bigdelim}
\usepackage{placeins}
\usepackage{breqn}
\usepackage{pifont}
\usepackage{booktabs}
\usepackage{subcaption}
\usepackage{hyperref}
\usepackage{authblk}

\title{The double traveling salesman problem with partial last-in-first-out loading constraints}

\author[1,2]{Jonatas B. C. Chagas\thanks{jonatas.chagas@iceb.ufop.br}}
\author[1]{Túlio A. M. Toffolo\thanks{tulio@toffolo.com.br}}
\author[1]{\\Marcone J. F. Souza\thanks{marcone@iceb.ufop.br }}
\author[3]{Manuel Iori\thanks{manuel.iori@unimore.it}}
\affil[1]{\mbox{Departamento de Computa\c{c}\~{a}o, Universidade Federal de Ouro Preto, Ouro Preto, Brazil}}
\affil[2]{Departamento de Inform\'{a}tica, Universidade Federal de Vi\c{c}osa, Vi\c{c}osa, Brazil}
\affil[3]{DISMI, Università degli studi di Modena e Reggio Emilia, Reggio Emilia, Italy}

\date{Technical Report \\[6mm] September 2020}

\begin{document}

\maketitle

\begin{abstract}
In this paper, we introduce the Double Traveling Salesman Problem with Partial Last-In-First-Out Loading Constraints (DTSPPL). It is a pickup-and-delivery single-vehicle routing problem, where all pickup operations must be performed before any delivery one because the pickup and delivery areas are geographically separated. The vehicle collects items in the pickup area and loads them into its container, a horizontal stack. After performing all pickup operations, the vehicle begins delivering the items in the delivery area. Loading and unloading operations must obey a partial Last-In-First-Out (LIFO) policy, i.e., a version of the LIFO policy that may be violated within a given reloading depth. The objective of the DTSPPL is to minimize the total cost, which involves the total distance traveled by the vehicle and the number of items that are unloaded and then reloaded due to violations of the standard LIFO policy. We formally describe the DTSPPL through two Integer Linear Programming (ILP) formulations and propose a heuristic algorithm based on the Biased Random-Key Genetic Algorithm (BRKGA) to find high-quality solutions. The performance of the proposed solution approaches is assessed over a broad set of instances. Computational results have shown that both ILP formulations have been able to solve only the smaller instances, whereas the BRKGA obtained good quality solutions for almost all instances, requiring short computational times.
\end{abstract}

\clearpage
\section{Introduction}
\label{sec:introduction}

Routing problems have been widely studied due to their practical and theoretical relevance. Studies in recent years have shown that transportation logistics is not only an essential part of the distribution of goods but also a vital field in the increasingly globalized world economy \citep{kherbash2015review}.

Among the routing problems, stands out the classical and well-known Traveling Salesman Problem (TSP). This problem aims to find the shortest Hamiltonian cycle in a weighted graph, and it is an $\mathcal{NP}$-Hard problem. Nevertheless, many studies devoted to its solution and advances in the combinatorial optimization area made it possible to solve large problem instances \citep{applegate2006traveling}.

Many extensions and variations of the TSP have been introduced and solved by the scientific community. Among these, we mention two classes of problems that are particularly interesting for our work. In \emph{Pickup and Delivery Problems}, the vehicle(s) transport(s) demands that have given origins and destinations \citep{BCI14, DS14}. In \emph{Routing Problems with Loading Constraints}, the route(s) performed by the vehicle(s) must be compatible with some loading policy \citep{iori2010routing}. These classes of problems have recently attracted the interest of many researchers because they are often noticed in the real world and also because of their combinatorial complexity. 

A problem that combines both pickups, deliveries, and loading is the Pickup and Delivery Traveling Salesman Problem with Last-In-First-Out Loading (PDTSPL). The PDTSPL arises in the context of a single vehicle, with a single access point (usually in the rear) for loading and unloading the transported items that must serve a set of customers and in such a way that pickups and deliveries obey the Last-In-First-Out (LIFO) policy. 

More specifically, in the PDTSPL each customer has a pickup location where the items are loaded into the loading compartment (a horizontal stack) of the vehicle, and a delivery location, where the collected items are delivered. At a pickup location, items are loaded and stored on top of the stack. At a delivery location, only the item located on top of the stack is available for unloading and delivery. The PDTSPL aims to find the minimum cost tour to visit all pickup and delivery locations while ensuring that the LIFO policy is obeyed.

The PDTSPL is applicable in situations where items are loaded and unloaded from the rear of the vehicle and where rearrangements among items are forbidden, because, e.g., the items may be large, heavy and/or fragile or dangerous to handle. By rearrangement, we mean the operation of temporarily dropping an already loaded item from the vehicle, and then reload it back, possibly in a different position.

According to \cite{iori2010routing}, the first work related to the PDTSPL was carried out by \cite{ladany1984optimal}. Posteriorly, the PDTSPL has also been studied by other researchers. \cite{carrabs2007additive} and \cite{cordeau2010branch} presented exact algorithms, whilst \cite{carrabs2007variable} and \cite{li2011tree} proposed heuristics.

Several other studies address pickup and delivery problems with LIFO loading constraints in scenarios where rearrangement operations are not allowed. Among these, we cite the Pickup and Delivery Traveling Salesman Problem with Multiple Stacks (PDTSPMS) \citep{cote2009branch} and the Double Traveling Salesman Problem with Multiple Stacks (DTSPMS) \citep{petersen2009double}. 

The PDTSPMS is a variant of the PDTSPL where the vehicle has its loading compartment divided into multiple horizontal stacks. On each stack, the loading and unloading operations must obey the LIFO principle. However, each stack is independent, so that there are no mutual constraints among the stacks. In this way, there is an increase in the flexibility of loading/unloading operations, which can lead to a reduction in transportation costs. The PDTSPMS was mainly treated by exact algorithms that employ different branch-and-cut schemes \citep{cote2009branch, sampaio2016new, pereira2018formulations}.

\cite{petersen2009double} introduced the DTSPMS from a real-world case that a software company encountered with one of its customers. The DTSPMS is a particular case of the PDTSPMS in which the pickup and delivery operations are completely separate because the pickup and delivery areas are geographically distant. Specifically, in the DTSPMS the vehicle starts its route in a pickup depot, performs all pickup operations in its stacks according to the LIFO principle, then returns to the pickup depot, from where the container is transferred by ship, airplane, train, or by similar transport form to a delivery depot located in the delivery area. From there, all delivery operations must be performed according to the LIFO principle. The objective of the DTSPMS is to find two routes, one in the pickup area and the other in the delivery area, so that the sum of the distances traveled in both areas is the minimum possible, while ensuring the feasibility of the vehicle loading plan. The transportation cost between the two regions is not considered as part of the optimization problem.

Since \cite{petersen2009double} presented it, the DTSPMS has aroused great attention in the academic community, with several exact and heuristic approaches developed for its solution. \cite{felipe2009double} proposed a heuristic approach based on the Variable Neighborhood Search (VNS) metaheuristic. \cite{petersen2010exact} proposed different exact mathematical formulations, including a branch-and-cut algorithm. \cite{lusby2010exact} presented an exact method based on matching $k$-best tours for each of the regions separately. \cite{carrabs2010branch} developed a branch-and-bound algorithm for the Double Traveling Salesman Problem with Two Stacks (DTSP2S), a particular case of the DTSPMS in which the vehicle has exactly two stacks. \cite{casazza2012efficient} studied the theoretical properties of the DTSPMS, analyzing the structure of DTSPMS solutions in two separate components: routes and loading plan. \cite{alba2013branch} proposed improvements to the branch-and-cut algorithm developed by \cite{petersen2010exact}, adding new valid inequalities that increased efficiency. \cite{urrutia2015dynamic} proposed a two-stage heuristic that first uses a local search algorithm to explore loading plan solutions, and then invokes a dynamic programming to construct optimal tours for each loading plan. \cite{barbato2016polyhedral} developed an exact algorithm that was able to solve instances involving containers of two stacks that were not previously solved in the literature. We refer to \cite{iori2015exact}, \cite{silveira2015heuristic}, \cite{chagas2016simulated} and \cite{chagas2019variable} for details upon the Double Vehicle Routing Problem with Multiple Stacks (DVRPMS), a variant of the DTSPMS that considers multiple vehicles.

A variant of the DTSPMS was suggested in Petersen's Ph.D. thesis \citep{petersen2009decision} (see page $68$, Section $2.5.2.4$), which we have named as the Double Traveling Salesman Problem with Multiple Stack and Partial Last-In-First-Out Loading Constraints (DTSPMSPL). The DTSPMSPL, like the DTSPMS, arises in transportation companies responsible for transporting large and fragile items from a pickup area to a delivery area, where these two areas are widely separated. However, the DTSPMSPL is more general as rearrangement operations are allowed as long as they obey a partial LIFO policy, that is, a version of the LIFO policy that may be violated within a given reloading depth. As stated by \cite{petersen2009decision}, the reason for a reloading depth is that replacing all items stored in the vehicle may be impractical due to the handling cost and the limited space available during reloading, so only a certain number of items may be placed outside the vehicle at any time. Thus, only the first $L$ items from the top of each stack may be relocated at any time. Note that rearrangement operations allow constructing shorter tours than those in which the classical LIFO policy must be respected. Although the base operation cost of the DTSPMSPL is the total routing cost, an additional handling cost should be paid for each item rearranged. Therefore, as stated by \cite{petersen2009decision}, partial LIFO constraints allow posing the question of what price transportation companies would be willing to pay for the opportunity to move one item.

The objective of the DTSPMSPL is to find a route in each area in such a way that the total cost is a minimum, and there exists a feasible loading/unloading plan following the partial LIFO policy. The total cost involves the routing cost, i.e., the sum of traveled distances in both areas, and the number of reloading operations performed in the loading/unloading plan, which have their cost is given in terms of the routing cost.

To the best of our knowledge, no study to date investigates the DTSPMSPL. Thus, in this work, we start this investigation, addressing a particular case of the DTSPMSPL, where the vehicle has its loading compartment as a single horizontal stack. We have named this new transportation problem as the Double Traveling Salesman Problem with Partial Last-In-First-Out Loading Constraints (DTSPPL).

\begin{figure*}[!ht]
	\centering
	\includegraphics[scale=0.26]{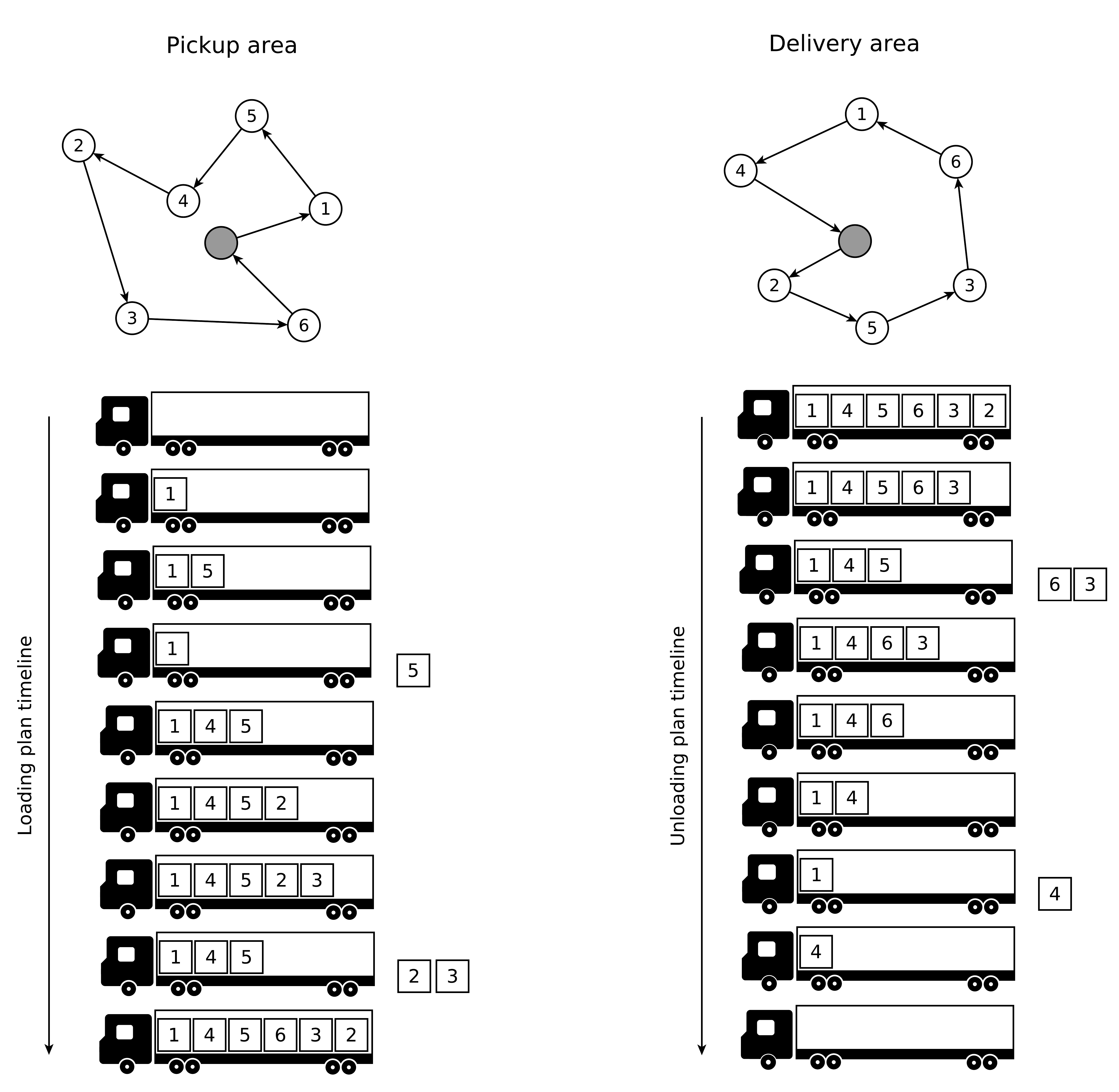}
	\caption{Solution of a DTSPPL instance involving 6 customers and reloading depth 2.}
	\label{fig:dtspplifo_example}
\end{figure*}

In order to clarify the characteristics of the DTSPMSPL/DTSPPL, we depict in Figure~\ref{fig:dtspplifo_example} a solution example of an instance with $6$ customers, considering a reloading depth equal to $2$, that is, $L = 2$. The vehicle starts its pickup route from the pickup depot (gray vertex on the top left part of the figure). It travels to the pickup position of customer $1$, storing its item in the stack. Next, it visits the pickup position of customer $5$, storing its item on the top of the horizontal stack. Then, it travels to pickup position customer $4$. At this point, a rearrangement is performed: the item of customer $5$ is removed from the stack, then the item of customer $4$ is placed in the stack and finally the item of customer $5$ is replaced into the stack. The vehicle continues its pickup route as shown in the figure until all items have been collected and stored in the stack, then the vehicle returns to the pickup depot. Notice that the reloading sequence may be in any order; thus, items do not need to remain in the same relative positions before their rearrangements, as occurs when the vehicle visits the pickup location of customer $6$. Upon arrival at the pickup depot, the container is transferred to the depot (gray vertex on the top right part of the figure) located in the delivery area, from where it is again transferred to a vehicle that then executes the delivery operations. In our example, first, the vehicle travels to the delivery position of customer $2$ without the requirement of any rearrangement. Next, it travels to the delivery position of customer $5$, where items $6$ and $3$ need to be removed before delivering item $5$. The delivery operations continue, as shown in the figure, until all customers are served. In the end, the vehicle returns to the delivery depot. Note that in this example, there were $3$ rearrangements in the pickup area (loading plan) and $3$ ones in the delivery area (unloading plan), totaling $6$ rearrangements.

\begin{figure}[!ht]
	\centering
	\includegraphics[scale=0.28]{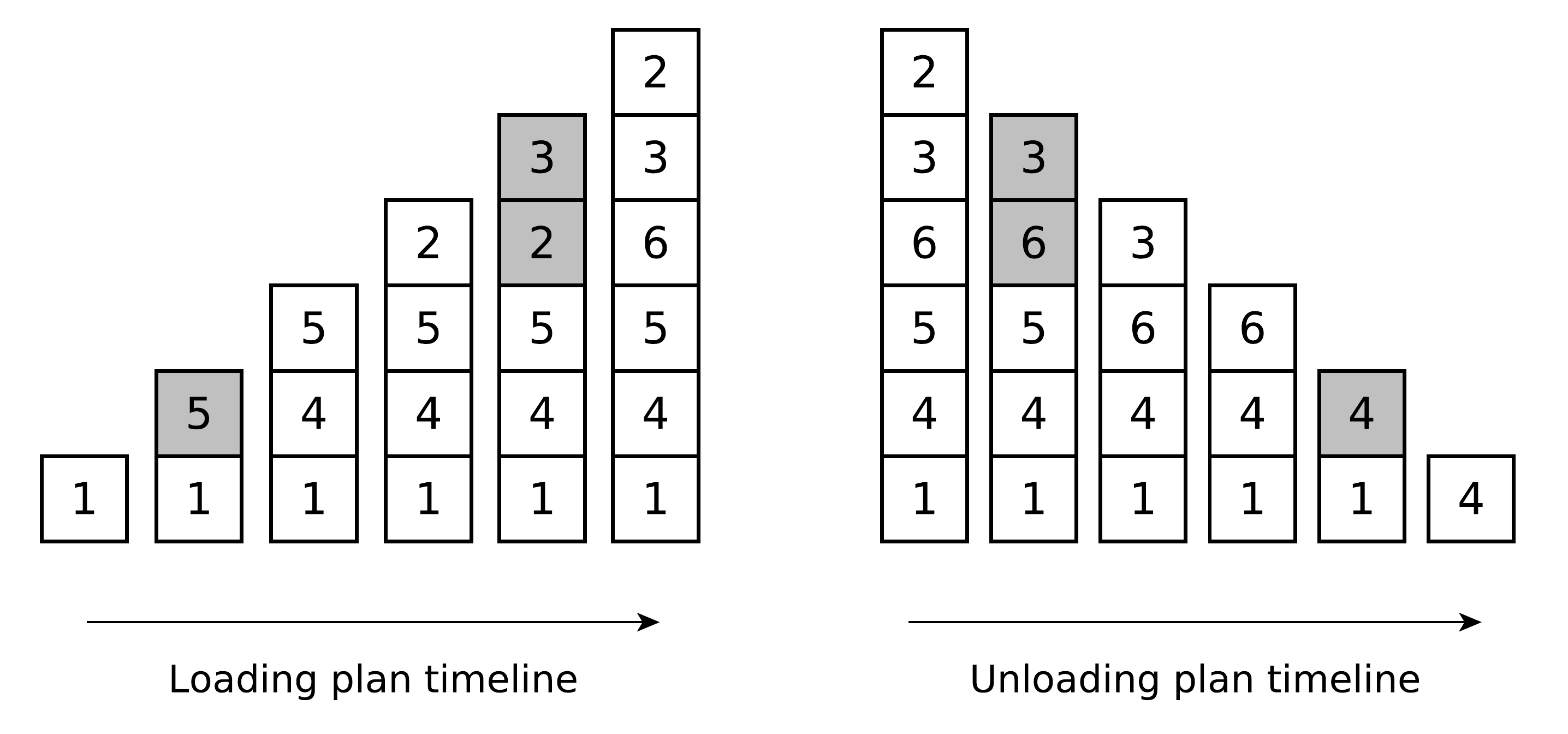}
	\caption{A practical representation of loading and unloading plans of the solution shown in Figure~\ref{fig:dtspplifo_example}.}
	\label{fig:dtspplifo_example_container}
\end{figure}

Figure~\ref{fig:dtspplifo_example_container} illustrates in a practical way the loading and unloading plan of the solution described in Figure~\ref{fig:dtspplifo_example}. Note that each column in Figure~\ref{fig:dtspplifo_example_container} indicates the container configuration after each pickup or delivery operation. Note also that it is possible to determine which items have been relocated (highlighted in gray) and how they have been reloaded by analyzing adjacent pairs of container configurations.

No previous work has approached the DTSPMSPL/DTSPPL. Nonetheless, \cite{ladany1984optimal} addressed a similar problem in which the reshuffling of all goods inside a container is allowed and causes costs and time losses. They investigated a real-world scenario in which identically sized crates should be transported from metropolitan area $A$ to metropolitan area $B$ using a single vehicle. The authors have solved small-size instances exactly by using an enumeration procedure. In addition, \cite{veenstra2017pickup} also approached a similar problem, named Pickup and Delivery Traveling Salesman Problem with Handling Costs (PDTSPH). It is a variant of the PDTSPL where rearrangement operations are allowed only at delivery locations, and, as in the problem studied by \cite{ladany1984optimal}, there is no maximum depth for reloading, i.e., at any delivery location, all items stored in the container may be relocated. The authors proposed a binary integer program for the PDTSPH, considering that the reloading sequence is the inverse of the unloading sequence, i.e., the items remain in the same relative positions before their rearrangements. They have also developed a Large Neighborhood Search (LNS) heuristic, which considers the reloading policy adopted in the binary integer program and another one where the reloaded items are positioned in the sequence in which they will be delivered. Their results show that this last reloading policy reduces the number of rearrangement operations. 

It is important to stress that by allowing rearrangement operations in both regions, a smaller reloading depth may be needed to rearrangement items according to given a pair of tours $\pi^P$ and $\pi^D$ (pickup and delivery tours, respectively). To illustrate this, consider the two different pairs of tours in Figure~\ref{fig:scenarios}. In the first pair (scenario \#1), we exemplify what might happen when rearrangements are allowed only in the pickup region. Note that the item of the $n$-th customer is the last to be collected and also the last to be delivered. Therefore, a reloading depth equal to $n-1$ is needed to allow unloading all $n-1$ items before collecting the $n$-th item in the pickup region, and then store it in the first position of the container, thus preparing the container for all deliveries without any rearrangement operation. In turn, if rearrangements are allowed only in the delivery region, a reloading depth equal to $n-1$ to construct feasible loading and unloading plans from $\pi^P$ and $\pi^D$ is also needed in the second pair of tours (scenario \#2). This reloading depth is needed since the first item to be delivered is stored in the first position of the container during collection.

\begin{figure}[!ht]
    \centering
	$\underbrace{\begin{array}{l}\pi^P = \langle\,0 \rightarrow 1 \rightarrow 2 \rightarrow ... \rightarrow n-1 \rightarrow n \rightarrow 0\,\rangle \\\pi^D = \langle\,0 \rightarrow n-1 \rightarrow n-2 \rightarrow ... \rightarrow n \rightarrow 0\,\rangle \end{array}}_{\text{\large scenario \#1}}\quad\underbrace{\begin{array}{l}\pi^P = \langle\,0 \rightarrow 1 \rightarrow 2 \rightarrow ... \rightarrow n-1 \rightarrow n \rightarrow 0\,\rangle \\\pi^D = \langle\,0 \rightarrow 1 \rightarrow n \rightarrow n-1 \rightarrow ... \rightarrow 2 \rightarrow 0\,\rangle \end{array}}_{\text{\large scenario \#2}}$
	\caption{Scenarios that justify the importance of allowing rearrangement operations in both regions.}
	\label{fig:scenarios}
\end{figure}

In both scenarios showed in Figure~\ref{fig:scenarios}, when rearrangement operations are allowed in the pickup and delivery regions, a reloading depth equal to 1 is enough to construct feasible loading and unloading plans. Therefore, in the context of the DTSPMSPL/DTSPPL, where rearrangement many items may be impractical due to the handling cost and the limited space available during reloading, it is crucial to allow rearrangement operations in both regions. Note that this can be done without any additional resources regarding those already available in the DTSPMS context. The same equipment used to load/unload an item can be used to unload and reload other items at each pickup and delivery point.

As is commonly addressed in the literature, we do not allow rearrangement operations at the depot. However, we can formulate scenarios in which rearrangements are also allowed at the depot by considering a fictitious item localized at the depot. If rearrangements at the depot are interesting, the fictitious item will be used to do them. It must be stressed that, in this approach, the reloading depth at the depot is limited to the same one used on the routes.

In the remainder of this article, we present our contributions. In Section~\ref{sec:problem-description}, we formally describe the DTSPPL via two Integer Linear Programming (ILP) formulations. Section~\ref{sec:brkga} describes a heuristic algorithm based on the concept of the Biased Random-Key Genetic Algorithm (BRKGA), which is able to find high-quality solutions for the DTSPPL in shorter computational time. Section~\ref{sec:computational-experiments} reports the experiments and analyzes the performance of the proposed solution approaches. Finally, in Section~\ref{sec:conclusions}, we present the conclusions and give suggestions for further investigations.

\section{Problem description and mathematical formulations}
\label{sec:problem-description}

In this section, we present the necessary notation to mathematically describe the DTSPPL and then propose two compact ILP formulations. Our mathematical formulations differ from each other in the way that route constraints are imposed. In both formulations, it is needed to provide for the other constraints of the models the visiting order of the customers in order to construct the loading and unloading plans.

An alternative formulation to those described as follows in this section could be designed based on infeasible path constraints. In this formulation, the problem should be decomposed into its routing and loading/unloading components. The routing component goal would be to construct a tour for each region, while loading/unloading component would aim at solving a packing problem from the fixed pair of tours found by the routing component. The packing goal would be to construct a complete feasible solution for the problem or identify cuts that eliminate tours that do not allow construction feasible loading and unloading plans. This strategy was used, e.g., for solving the DTSPMS in a branch-and-cut algorithm proposed by~\cite{alba2013branch}. As stated by~\cite{alba2013branch}, in the DTSPMS context, from a given pair of tours, it is possible to construct a precedence graph, and then from it determining whether the tours are compatible with the LIFO constraints. Their separation algorithms have been developed from this property in order to find cuts for their branch-and-cut method. Note that our packing problem is much more complex. Due to partial LIFO constraints, we may not construct a precedence graph and work on it once items may be rearranged. Therefore, we cannot efficiently solve the DTSPPL by using this strategy.

\subsection{Problem description}

The DTSPPL can be formally described as follows. Let $C = \{1, 2, ..., n\}$ be the set of $n$ customer requests, $V_{c}^{P} = \{1^{P}, 2^{P}, ..., n^{P}\}$ the set of pickup locations and $V_{c}^{D} = \{1^{D}, 2^{D}, ..., n^{D}\}$ the set of delivery locations. For each customer request $i \in C$, an item has to be transported from the pickup location $i^{P}$ to the delivery location $i^{D}$.

The DTSPPL is defined on two complete directed graphs, $G^{P} = (V^{P}, A^{P})$ and $G^{D} = (V^{D}, A^{D})$, which represent the pickup and delivery areas, respectively. The sets $V^{P} = \{ 0^{P} \} \;\cup\; V_{c}^{P}$ and $V^{D} = \{ 0^{D} \} \;\cup\; V_{c}^{D}$ represent the vertices in each area, with $0^{P}$ and $0^{D}$ denoting the depots of the pickup and delivery areas, respectively. The sets of arcs in the pickup and delivery areas are defined by $A^{P} = \{(i^{P}, j^{P}, c_{ij}^{P}) \;\; \forall i^{P} \in V^{P}, \forall j^{P} \in V^{P} \;\vert\; j^{P} \neq i^{P}\}$ and $A^{D} = \{(i^{D}, j^{D}, c_{ij}^{D}) \;\; \forall i^{D} \in V^{D}, \forall j^{D} \in V^{D} \;\vert\; j^{D} \neq i^{D}\}$, where $c_{ij}^{P}$ and $c_{ij}^{D}$ correspond to the travel distances associated with arcs $(i^{P}, j^{P})$ and $(i^{D}, j^{D})$, respectively. For convenience of notation, when no confusion arises we also refer to sets $V_{c}^{P}$ and $V_{c}^{D}$ as the set of requests $C$, and we use $i$ to denote both $i^{P}$ and $i^{D}$ and $(i, j)$ to denote both $(i^{P}, j^{P})$ and $(i^{D}, j^{D})$.

A feasible solution $s$ for the DTSPPL consists of a Hamiltonian cycle on a graph $G^{P}$ that starts at the pickup depot $0^{P}$, another Hamiltonian cycle on graph $G^{D}$ that begins at the delivery depot $0^{D}$, and a loading/unloading plan. Besides, the two Hamiltonian cycles and the loading and unloading plan must obey the partial LIFO policy, which is defined by the maximum reloading depth $L$.

Let us denote by $\mathcal{F}$ the set of all feasible solutions for a DTSPPL instance. Each solution $s \in \mathcal{F}$ has a cost $c_{s}$ that involves the travel distance on the two Hamiltonian cycles and the number of rearrangements performed on the loading and unloading plan, with a cost $h$ associated with a single item rearrangement. The objective of the DTSPPL is to find a solution $s^{*} \in \mathcal{F}$, so that $c_{s^{*}} = \min_{s \in \mathcal{F}} c_{s}$.

\subsection{Integer linear programming formulation 1}

To better explain the first proposed ILP formulation (ILP1) for the DTSPPL, we categorize the constraints under three groups: ($i$) routes structuring constraints, ($ii$) loading/unloading plan and partial LIFO constraints, and ($iii$) reloading control constraints. After describing all constraints, we present the objective function of the DTSPPL.

Throughout mathematical modeling, we also use the notation $[a,\,b]$ to denote the set $\{a, a + 1, ..., b-1, b\}$. Note that for any $a > b$, $[a,\,b]$ is an empty set.

\subsubsection{Route structuring constraints}

In order to characterize the pickup and delivery routes, we define constraints \eqref{eq:ilp1-01}-\eqref{eq:ilp1-07}, which use binary decision variables $x_{ij}^{kr},\;\forall r \in \{P, D\},\, k \in [1,n+1],\, (i, j) \in A^{r}$, that assume value $1$ if arc $(i, j)$ is the $k$-th traveled arc by the vehicle in the area $r$, and value $0$ otherwise.

\begin{align}
	\label{eq:ilp1-01}
	&\sum_{j\;:\;(0,\;j)\;\in\;A^{r}} x_{0j}^{1r} = 1 &\begin{array}{r}r \in \{P, D\}\end{array}\\
	\label{eq:ilp1-02}
	&\sum_{i\;:\;(i,\;0)\;\in\;A^{r}} x_{i0}^{n+1,r} = 1 &\begin{array}{r}r \in \{P, D\}\end{array}\\
	\label{eq:ilp1-03}
	&\sum_{(i,\;j)\;\in\;A^{r}} x_{ij}^{kr} = 1 &\begin{array}{r}r \in \{P, D\},\, k \in [2,n]\end{array}\\
	\label{eq:ilp1-04}
	&\sum_{k\;\in\;[1,\;n]} \sum_{\;\;i\;:\;(i,\;j)\;\in\;A^{r}} x_{ij}^{kr} = 1 &\begin{array}{r}r \in \{P, D\},\, j \in V_{c}^{r}\end{array}\\
	\label{eq:ilp1-05}
    &x_{ij}^{kr} \leq \sum_{i'\;:\;(i',\;i)\;\in\;A^{r}} x_{i'i}^{k-1,r} &\begin{array}{r}r \in \{P, D\},\, k \in [2,n+1],\, (i,j) \in A^{r}\end{array}\\
    \label{eq:ilp1-06}
    &x_{ij}^{kr} \in \{0, 1\} &\begin{array}{r}r \in \{P, D\},\, k \in [1,n+1],\, (i, j) \in A^{r}\end{array}
\end{align}

Constraints \eqref{eq:ilp1-01} and \eqref{eq:ilp1-02} force the vehicle to leave from the depot and return to it using, respectively, the first and the last arc in each area. Constraints \eqref{eq:ilp1-03} guarantee that only a single arc may be the $k$-th one of each route. Constraints \eqref{eq:ilp1-04} guarantee that every customer is served. Constraints \eqref{eq:ilp1-05} establish the flow conservation for each vertex, and constraints \eqref{eq:ilp1-06} define the domain of the decision variables used to represent the routes.

\subsubsection{Loading/unloading plan and partial LIFO constraints}

For the purpose of representing the loading plan, we define binary decision variables $y_{j\ell}^{kP},\;\forall k \in [1,n],\, \ell \in [1,k],\, j \in C$, that assume value $1$ if the item referring to the customer request $j$ is stored in position $l$ on the $k$-th container configuration in the pickup area, and value $0$ otherwise. We also define other binary decision variables $y_{j\ell}^{kD},\;\forall k \in [1,n],\, \ell \in [1,n-k+1],\, j \in C$, to represent the unloading plan, which has the same meaning as the previous variable but considers the delivery area. With these variables, we can ensure the feasibility of the loading and unloading plans throughout constraints \eqref{eq:ilp1-07}-\eqref{eq:ilp1-15}, which are next explained in detail.

\begin{align}
	\label{eq:ilp1-07}
    &\sum_{j\;\in\;C} y_{j\ell}^{kP} = 1 &\begin{array}{r}k \in [1,n],\, \ell \in [1,k]\end{array}\\
    \label{eq:ilp1-08}
	&\sum_{j\;\in\;C} y_{j\ell}^{kD} = 1 &\begin{array}{r}k \in [1,n],\, \ell \in [1,n-k+1]\end{array}\\
	\label{eq:ilp1-09}
   	&\sum_{l\;\in\;[1,\;k]} y_{j\ell}^{kP} = \sum_{k'\;\in\;[1,\;k]} \sum_{\;\;i\;:\;(i,\;j)\;\in\;A^{P}} x_{ij}^{k'P} &\begin{array}{r}k \in [1,n],\, j \in C\end{array}\\[2mm]
    \label{eq:ilp1-10}
    &\sum_{l\;\in\;[1,\;n-k+1]} y_{j\ell}^{kD} = \sum_{k'\;\in\;[k,\;n]} \sum_{\;\;i\;:\;(i,\;j)\;\in\;A^{D}} x_{ij}^{k'D} &\begin{array}{r}k \in [1,n],\, j \in C\end{array}
\end{align}

\begin{figure}[!ht]
	\centering
	\includegraphics[scale=0.28]{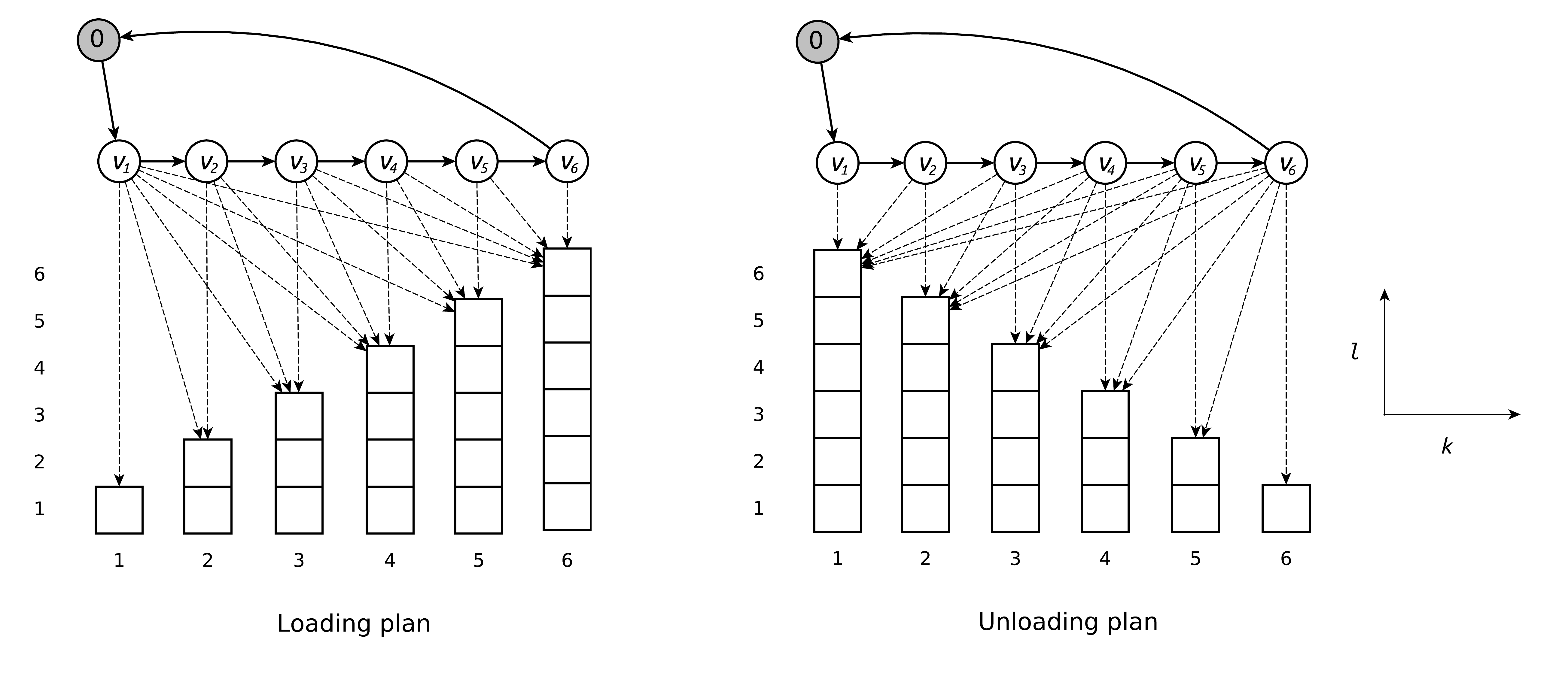}
    \caption{Graphical representation of constraints \eqref{eq:ilp1-09} and \eqref{eq:ilp1-10}.}
    \label{fig:loading_unloadoing_plan_fill_constraints}
\end{figure}

Constraints \eqref{eq:ilp1-07} and \eqref{eq:ilp1-08} ensure that only one item must occupy each container position in each of its configurations. Constraints \eqref{eq:ilp1-09} establish that in the pickup area the $k$-th container configuration has to contain all items collected at the vertices $v_{i},\;\forall i \leq k$. In turn, constraints \eqref{eq:ilp1-10} guarantee that in the delivery area the $k$-th container configuration has to contain all items that have not yet been delivered at the vertices $v_{i},\;\forall i \geq k$. Figure~\ref{fig:loading_unloadoing_plan_fill_constraints} depicts the operation of constraints \eqref{eq:ilp1-09} and \eqref{eq:ilp1-10} for an instance with 6 customer requests. Dashed arrows indicate which items must be in each container configuration according to the pickup and delivery tours, which are represented by the continuous lines that connect the vertices.

Constraints \eqref{eq:ilp1-11} certify that the first container configuration in the delivery area must be the same as the last container configuration in the pickup area. In other words, these constraints ensure that there is no rearrangement of items between the transfer from the pickup depot to the delivery depot. 

\begin{align}
    \label{eq:ilp1-11}
	&y_{j\ell}^{1D} = y_{j\ell}^{nP} &\begin{array}{r}\ell \in [1,n], j \in C\end{array}
\end{align}

\begin{figure}[!ht]
	\centering
	\includegraphics[scale=0.28]{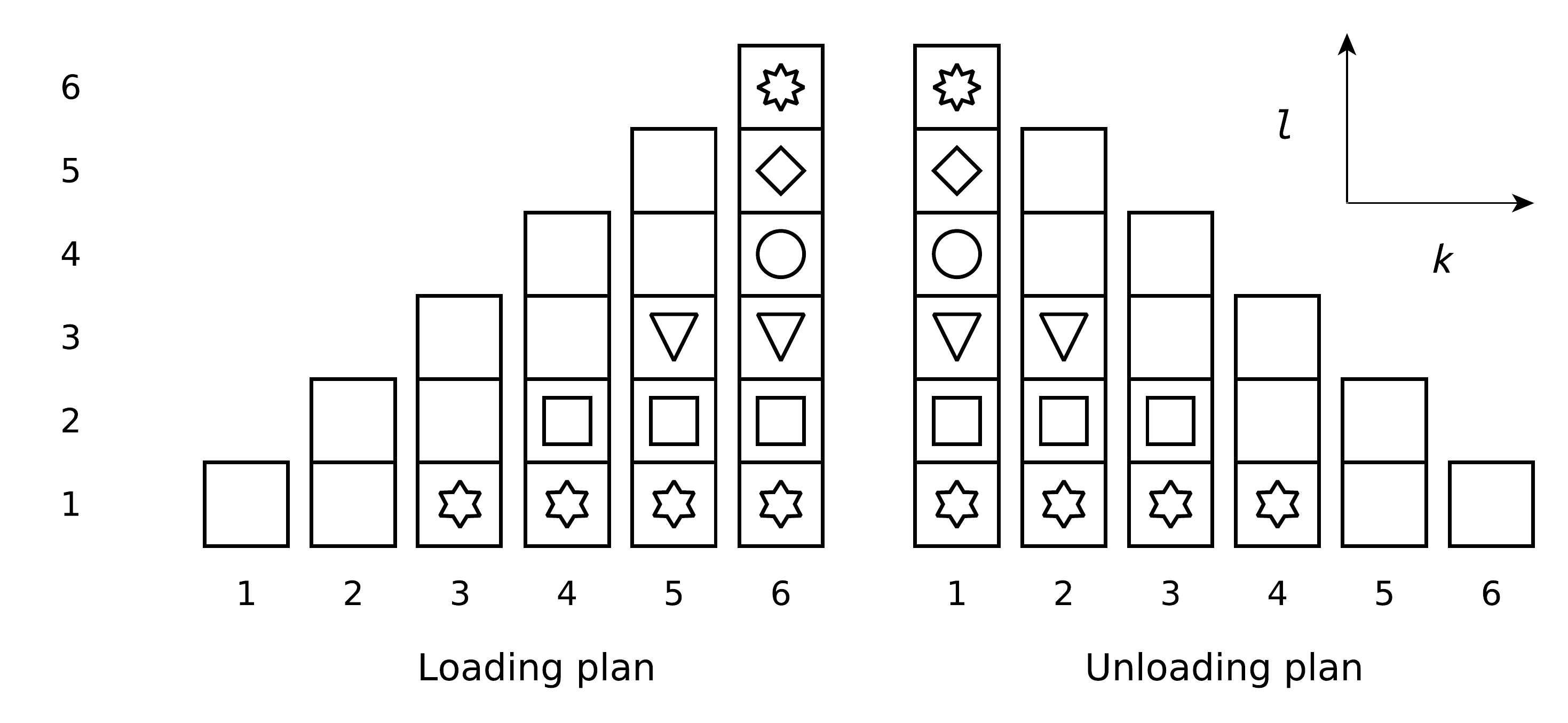}
	\caption{Graphical representation of constraints \eqref{eq:ilp1-11}-\eqref{eq:ilp1-13}.}
	\label{fig:loading_unloading_plan_reloading_depth_constraints}
\end{figure}

Figure~\ref{fig:loading_unloading_plan_reloading_depth_constraints} illustrates the  operation of constraints \eqref{eq:ilp1-11} for an instance with 6 customer requests. It also illustrates the operation of constraints \eqref{eq:ilp1-12} and \eqref{eq:ilp1-13} that ensure the loading plan obeys the partial LIFO policy, taking as example $L = 2$. Note that constraints \eqref{eq:ilp1-12} and \eqref{eq:ilp1-13} establish which items (represented by different geometric shapes) have to remain in their previous container positions in order not to violate the partial LIFO policy.

\begin{align}
    \label{eq:ilp1-12}
    &y_{j\ell}^{kP} = y_{j\ell}^{nP} &\begin{array}{r}k \in [1,n-1],\, \ell \in [1,k-L], j \in C\end{array}\\[2mm]
    \label{eq:ilp1-13}
    &y_{j\ell}^{kD} = y_{j\ell}^{1D} &\begin{array}{r}k \in [2,n],\, \ell \in [1,n-k-L+1],\, j \in C\end{array}
\end{align}

Finally, constraints \eqref{eq:ilp1-14} and \eqref{eq:ilp1-15} define the domain of the decision variables used to represent the loading and unloading plan.

\begin{align}
    \label{eq:ilp1-14}
    &y_{j\ell}^{kP} \in \{0, 1\} &\begin{array}{r}k \in [1,n],\, \ell \in [1,k],\, j \in C\end{array}\\        
    \label{eq:ilp1-15}
    &y_{j\ell}^{kD} \in \{0, 1\} &\begin{array}{r}k \in [1,n],\, \ell \in [1,n-k+1],\, j \in C\end{array}        
\end{align}

\subsubsection{Reloading control constraints}

\begin{figure}[!ht]
	\centering
	\includegraphics[scale=0.30]{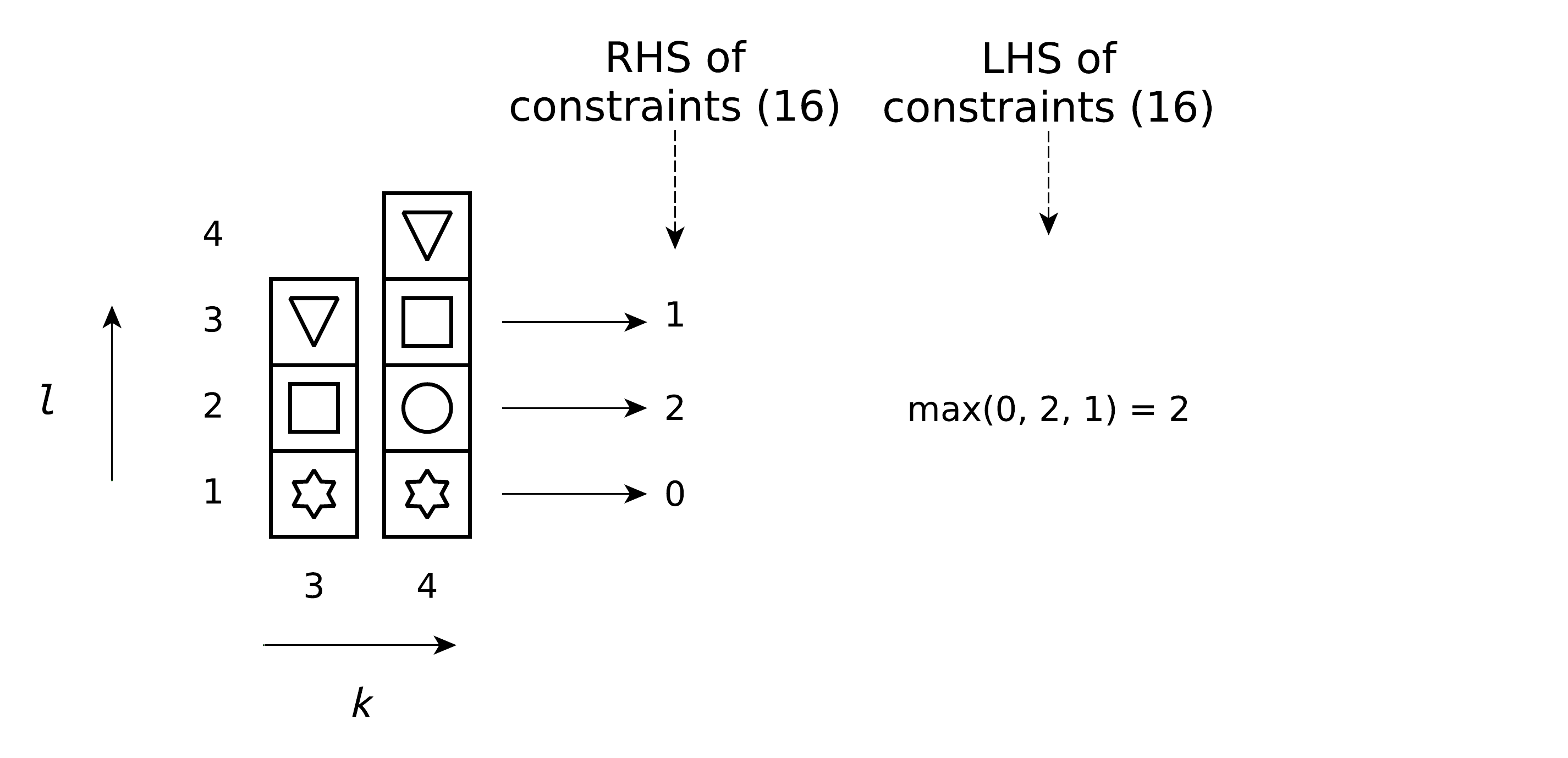}
	\caption{Graphical representation of constraints \eqref{eq:ilp1-16}.}
	\label{fig:reloading_control_constraints}
\end{figure}

To determine how many rearrangements are performed from the loading and unloading plans, we define decision variables $z^{kr},\;\forall r \in \{P, D\}, k \in [1,n-1]$, that indicate the number of rearrangements made in the $k$-th container configuration in area $r$. We also define constraints \eqref{eq:ilp1-16} and \eqref{eq:ilp1-17}, which are responsible for determining the number of rearrangements in the loading and unloading plans, respectively. Constraints~\eqref{eq:ilp1-16} analyze every pair of adjacent container configurations ($k$ and $k\text{+}1$-th) of the loading plan to determine the number of rearrangements made in the $k$-th container configuration according to the $k\text{+}1$-th configuration. Figure~\ref{fig:reloading_control_constraints} illustrates the operation of constraints~\eqref{eq:ilp1-16} by exemplifying the calculation of the number of rearrangements made in the $3$-th container configuration. In this example, the first ($\ell = 1$) item is not rearranged, as shown in the $4$-th container configuration; while the second ($\ell = 2$) and third ($\ell = 3$) items are. Note that the number of rearrangements is given from the deeper change in the $k$-th container configuration.
Constraints~\eqref{eq:ilp1-17} are similar to constraints~\eqref{eq:ilp1-16}, but they count the rearrangements made in the delivery area. Finally, constraints~\eqref{eq:ilp1-18} define the domain of these decision variables.

\begin{align} 
    \label{eq:ilp1-16}
    &z^{kP} \geq \Big(\,y_{j\ell}^{kP} - y_{j\ell}^{k+1,P}\,\Big)\cdot\Big(\,k-\ell+1\,\Big) &\begin{array}{r}k \in [1,n-1],\, \ell \in [1,k-1],\, j \in C\end{array}\\[2mm]
    \label{eq:ilp1-17}
	\begin{split}&z^{kD} \geq \Big(\,y_{j\ell}^{k+1,D} - y_{j\ell}^{kD}\,\Big)\cdot \Big(\,n-k-\ell+1\,\Big)\end{split} &\begin{array}{r}k \in [1,n-1],\, \ell \in [1,n-k],\, j \in C\end{array}\\[2mm]
	\label{eq:ilp1-18}
    &z^{kr} \in \mathbb{Z}^{+} &\begin{array}{r}r \in \{P, D\},\, k \in [1,n-1]\end{array}
\end{align}

\subsubsection{Objective function}

Constraints \eqref{eq:ilp1-01}-\eqref{eq:ilp1-18} are enough to represent all feasible solutions of the DTSPPL. Therefore, to complete the first mathematical model, we define the objective function \eqref{eq:ilp1-19}, which minimizes the total cost. It involves the distance traveled in both areas, as well as the cost of all rearrangements performed.

\begin{equation} \label{eq:ilp1-19}
	\min \sum_{r\,\in\,\{P,\,D\}} \sum_{(i,\,j)\,\in\,A^{r}} c_{ij}^{\,r} \;\cdot \sum_{k\,\in\,[1,\,n+1]} x_{ij}^{kr} \;+\; h \;\cdot \sum_{r\;\in\;\{P,\;D\}} \sum_{k\;\in\;[1,\;n-1]} z^{kr}
\end{equation}

\subsection{Integer linear programming formulation 2}

Our second ILP formulation (ILP2) uses binary decision variables $\chi_{ij}^{r},\;\forall r \in \{P, D\},\, (i, j) \in A^{r}$, to describe the vehicle route in each area. More specifically, each variable $\chi_{ij}^{r}$ assumes value $1$ if arc $(i, j)$ is traveled by the vehicle in area $r$, and value $0$ otherwise. Moreover, we use an integer variable $u_{j}^{r},\;\forall r \in \{P, D\},\, j \in [0, n]$, that gives the position of vertex $j$ in the route of area~$r$. With these new decision variables, we can use constraints \eqref{eq:ilp2-01}-\eqref{eq:ilp2-06}, instead of \eqref{eq:ilp1-01}-\eqref{eq:ilp1-06} adopted for ILP1, to define the routes of the vehicle. 

\begin{align}
	\label{eq:ilp2-01}
	&\sum_{j\;:\;(i,\;j)\;\in\;A^{r}} \chi_{ij}^{r} = 1 &\begin{array}{r}r \in \{P, D\},\, i \in V^{r}\end{array}\\
	\label{eq:ilp2-02}
	&\sum_{i\;:\;(i,\;j)\;\in\;A^{r}} \chi_{ij}^{r} = 1 &\begin{array}{r}r \in \{P, D\},\, j \in V^{r}\end{array}\\
	\label{eq:ilp2-03}
	&u_{j}^{r} \geq u_{i}^{r} + 1 - n \cdot \left(1 - \chi_{ij}^{r} \right) &\begin{array}{r}r \in \{P, D\},\, (i, j) \in A^{r}: j \neq 0\end{array}\\
	\label{eq:ilp2-04}
	&\chi_{ij}^{r} \in \{0, 1\} &\begin{array}{r}r \in \{P, D\},\, (i, j) \in A^{r}\end{array}\\
	\label{eq:ilp2-05}
	&u_{0}^{r} = 0 &\begin{array}{r}r \in \{P, D\}\end{array}\\
	\label{eq:ilp2-06}
	&u_{j}^{r} \in \{a \in \mathbb{Z}^{+}: a \leq n\} &\begin{array}{r}r \in \{P, D\},\, j \in C\end{array}
\end{align}

Constraints \eqref{eq:ilp2-01}, \eqref{eq:ilp2-02} and \eqref{eq:ilp2-04} ensure that each pickup and delivery location is visited exactly once, whilst constraints \eqref{eq:ilp2-03}, \eqref{eq:ilp2-05} and \eqref{eq:ilp2-06} impose the subcycle elimination.

To complete ILP2, we define constraints \eqref{eq:ilp2-07} and \eqref{eq:ilp2-08}, and the objective function \eqref{eq:ilp2-09}. Moreover, we also include constraints \eqref{eq:ilp1-07}, \eqref{eq:ilp1-08} and \eqref{eq:ilp1-11}-\eqref{eq:ilp1-18}, which have been previously defined for ILP1.

\begin{align}
    \label{eq:ilp2-07}
   	&1 \geq \sum_{l\;\in\;[1,\;k]} y_{j\ell}^{kP} \geq \frac{k - u_{j}^{P} + 1}{k} &\begin{array}{r}k \in [1,n],\, j \in C\end{array}\\
    \label{eq:ilp2-08}
    &1 \geq \sum_{l\;\in\;[1,\;n-k+1]} y_{j\ell}^{kD} \geq \frac{u_{j}^{D} - k + 1}{n - k + 1} &\begin{array}{r}k \in [1,n],\, j \in C\end{array}
\end{align}

Note that constraints \eqref{eq:ilp2-07} and \eqref{eq:ilp2-08} have the same aim as constraints \eqref{eq:ilp1-09} and \eqref{eq:ilp1-10}, which ensure the correct assignment of items to the loading compartment throughout the pickup and delivery routes. Note also that equation \eqref{eq:ilp2-09}, similarly to \eqref{eq:ilp1-19}, describes the total DTSPPL cost to be minimized.

\begin{equation} \label{eq:ilp2-09}	
	\min \sum_{r\,\in\,\{P,\,D\}} \sum_{(i,\,j)\,\in\,A^{r}} c_{ij}^{\,r} \;\cdot \chi_{ij}^{r} \;\;+\;\; h \;\cdot \sum_{r\;\in\;\{P,\;D\}} \sum_{k\;\in\;[1,\;n-1]} z^{kr}
\end{equation}

\subsection{Providing the ILP models with a feasible initial solution} \label{sec:ilp_overall}

We have solved models ILP1 and ILP2 using Gurobi Optimizer, which is currently one of the best ILP optimization solvers. However, we have noticed that, even for small instances, Gurobi had difficulties in solving both models within a reasonable time. Therefore, in order to help the optimization process, we compute a feasible DTSPPL solution and initialize both models with it. For the initial solution, we consider the case where rearrangements are not allowed (i.e., $L = 0$). Thus, as every loading/unloading operation must verify the classic LIFO principle, the pickup and delivery routes must be exactly opposite each other, since the vehicle has its loading compartment as a single stack. Therefore, we can solve the initial solution by solving a TSP instance on a graph where each arc $(i,\,j)$ is associated a cost $c_{ij} = c_{ij}^{\,P} + c_{ji}^{\,D}$. This strategy was also used by \cite{felipe2009double} to compute an initial solution for the DTSPMS. In this work, we solve a TSP instance via the classical two-index model for the TSP (see. e.g., \cite{gutin2006traveling}) by adding subtour elimination constraints iteratively until the incumbent solution does not contain subtours.

\section{A biased random-key genetic algorithm}
\label{sec:brkga}

In this section, we describe a heuristic algorithm based on the Biased Random-Key Genetic Algorithm (BRKGA) \citep{gonccalves2011biased}. Although metaheuristic algorithms based on local search such as Tabu Search, Variable Neighborhood Search and Iterated Local Search, among others (see, e.g.,  \cite{talbi2009metaheuristics}) are more often applied to address vehicle routing problems, we have chosen an evolutionary algorithm that works with an indirect representation of its individuals (where each individual represents a problem solution). The justification for this choice is that the DTSPPL has a high dependency on routes and loading/unloading plans. So, working with direct solutions may not be practicable, since defining efficient moves that can navigate between feasible solutions, and, especially, escape from infeasible solution space is extremely hard. In turn, an indirect representation of DTSPPL solutions allows us to navigate in the feasible solution space through simple genetic operators, quickly producing a high number of feasible solutions. This representation strategy has been successfully applied to several complex optimization problems in recent years \citep{gonccalves2012parallel, resende2012biased, gonccalves2013biased, lalla2014biased, gonccalves2015biased, santos2018thief}.

In the remainder of this section, we present the main components of the proposed BRKGA (Sections \ref{sec:brkga_encode_structure}-\ref{sec:brkga_next_generation}) and then describe how these components are combined together (Section \ref{sec:brkga_overall}).

\subsection{Encoding structure}
\label{sec:brkga_encode_structure}

BRKGAs, as well as classic Genetic Algorithms (GAs) (see \cite{mitchell1998introduction} for a reference), are evolutionary metaheuristics that mimic the processes of Darwinian Evolution. Basically, a GA maintains a population of individuals, each encoding a solution to the problem at hand. Through the use of stochastic evolutionary processes (selection, recombination, and diversification) over the population, individuals with higher fitness tend to survive, thus guiding the algorithm to explore more promising regions of the solutions space.

Each individual in BRKGAs is represented by a vector of random-keys, i.e., a vector of real numbers that assume values in the continuous interval [0, 1]. This representation is generic because it is independent of the problem addressed. Therefore, a deterministic procedure (to be presented later) is necessary to decode each individual (a vector of random-keys) to a feasible solution of the problem at hand (the DTSPPL in our case).

\begin{figure}[!ht]
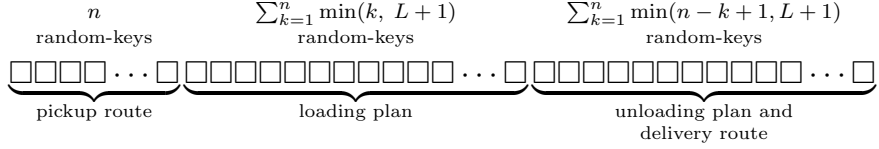

	\centering
	\begin{equation}
	    \qquad \qquad \quad \large\underbrace{\square \square \square \square \cdots \square}_{\text{pickup route}}^{\footnotesize \begin{array}{c}n\\ \text{\scriptsize random-keys}\\[1mm]\end{array}}\underbrace{\square \square \square \square \square \square \square \square \square \square \square \cdots \square}_{\text{\scriptsize loading plan}}^{\footnotesize \begin{array}{c}\sum_{k=1}^{n} \min(k,\; L+1)\\\text{\scriptsize random-keys}\\[1mm]\end{array}}\underbrace{\square \square \square \square \square \square \square \square \square \square \square \cdots \square}_{\text{\parbox{80pt}{\centering \scriptsize unloading plan and delivery route}}}^{\footnotesize \begin{array}{c}\sum_{k=1}^{n} \min(n-k+1, L+1)\\\text{\scriptsize random-keys}\\[1mm]\end{array}} \nonumber
	\end{equation}
	\caption{Chromosome structure.}
	\label{fig:brkga_encode_structure}
\end{figure}

In Figure~\ref{fig:brkga_encode_structure}, we show the structure defined to represent each BRKGA individual for the DTSPPL. We divide this structure into three partitions: pickup route, loading plan, and unloading plan and delivery route. The first one consists of $n$ random-keys, which are responsible for determining the pickup route. The next $\sum_{k=1}^{n} \min(k,\; L+1)$ random-keys determine the loading plan carried out along the pickup route. Finally, the last $\sum_{k=1}^{n} \min(n-k+1, L+1)$ random-keys define the entire unloading plan and also the delivery route. 

\subsection{Decoding procedure}
\label{sec:brkga_decoding_procedure}

As stated before, each individual $p \in \mathcal{P}$ has a generic representation in a BRKGA. Therefore, to determine the fitness of $p$, we developed a procedure that generates a feasible solution $s$ from $p$. The fitness of $p$ is then defined proportionally to the quality of~$s$.

\begin{figure*}[!ht]
	\centering
	\includegraphics[width=\textwidth]{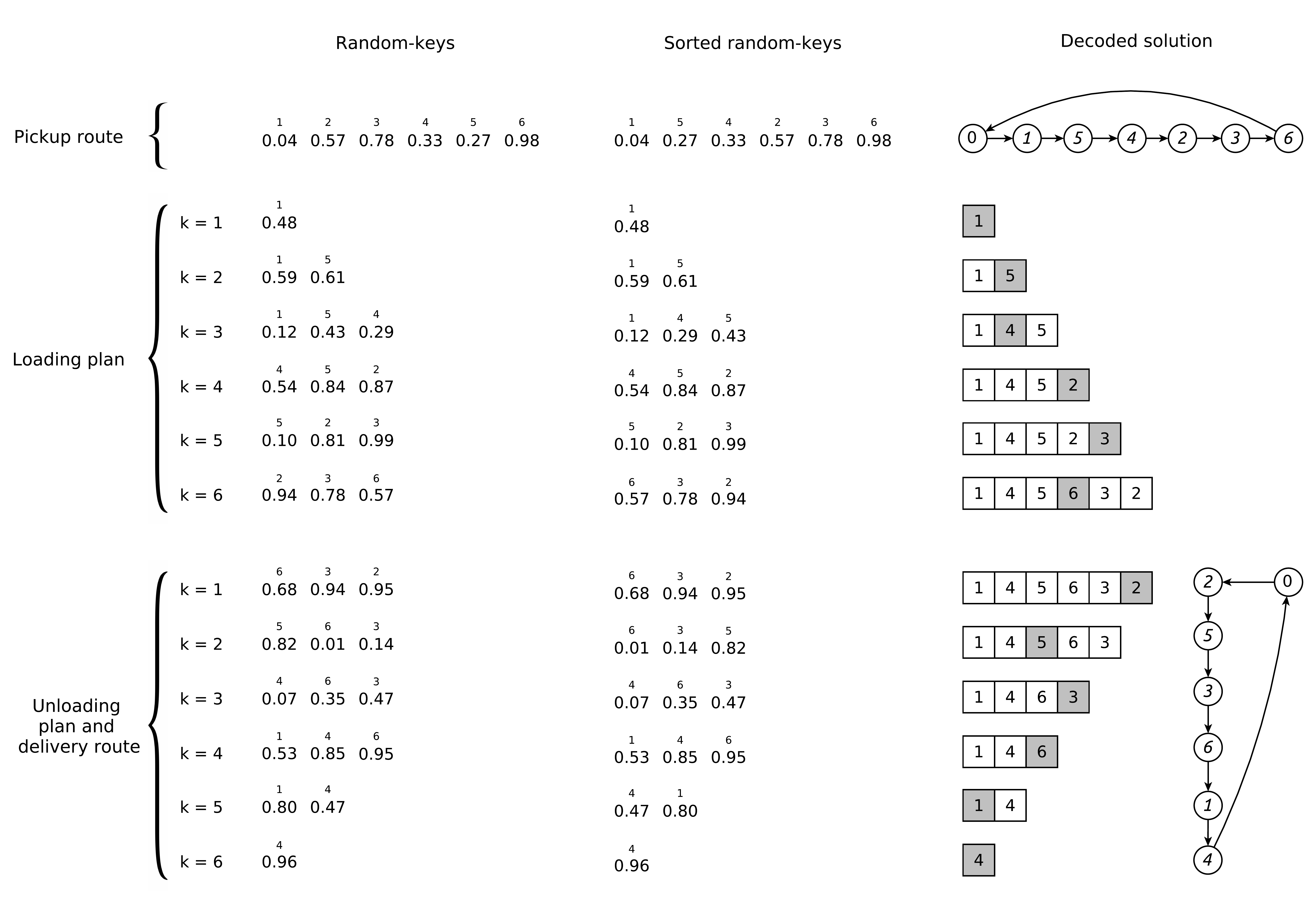}
	\caption{Decoding of the vector of random-keys $\langle$ {\footnotesize 0.04, 0.57, 0.78, 0.33, 0.27, 0.98, 0.48, 0.59, 0.61, 0.12, 0.43, 0.29, 0.54, 0.84, 0.87, 0.10, 0.81, 0.99, 0.94, 0.78, 0.57, 0.68, 0.94, 0.78, 0.57, 0.68, 0.94, 0.78, 0.57, 0.68, 0.94, 0.95, 0.82, 0.01, 0.14, 0.07, 0.35, 0.47, 0.53, 0.85, 0.95, 0.80, 0.47, 0.96} $\rangle$ to the solution shown in Figure~\ref{fig:dtspplifo_example}.}
	\label{fig:brkga_decoding_example}
\end{figure*}

Our decoding procedure consists of three stages, which must be sequentially performed due to the dependency among them. Figure~\ref{fig:brkga_decoding_example} depicts how these stages are performed in order to decode the solution shown in Figure~\ref{fig:dtspplifo_example} from a vector of random-keys.
Initially, the $n$ pickup locations are mapped on the first $n$ random-keys. Then, we sort the pickup locations according to the values of the mapped random-keys. The sorted pickup locations define the pickup route $\pi^{P}$ performed by the vehicle. In Figure~\ref{fig:dtspplifo_example}, the first $n$ random-keys produce the pickup route $\pi^{P} = \langle 0, 1, 5, 4, 2, 3, 6, 0 \rangle$.

We characterize the loading plan from $\pi^{P}$ and the next $\sum_{k=1}^{n} \min(k,\; L+1)$ random-keys. The loading plan is created iteratively in $n$ steps, where each of them depends on the previous one. Iteratively, for each $k \in \{1, 2, ..., n\}$, we map the $\min(k,\;L+1)$ items that may be relocated without violating the partial LIFO constraints to $\min(k,\;L+1)$ random-keys. Next, we sort the items in non-decreasing order according to the values of the mapped random-keys to define the $k$-th container configuration in the pickup area. For the sake of clarity, consider the steps shown in Figure~\ref{fig:brkga_decoding_example}. At first, the container is empty, so we just store in it the item (highlighted in gray) of the first customer visited according to $\pi^{P}$. Notice that, though unnecessary, we kept a random-key (0.48 in this example) to decode the first operation of the loading plan. We have decided to keep it for simplicity, to follow the same pattern used for the other loading operations. To define the second container configuration, we map item 1 (it may be relocated from the previous container configuration) to the random-key 0.59 and item 5 (the second customer visited according to $\pi^P$) to the random-key 0.61. After sorting these items according to the values of the mapped random-keys, items 1 and 5 are stored in the container following the sorted order. The decoding process continues by mapping the items that may be relocated at each time to the random-keys, and then sort them to define each remaining pickup container configuration. 

Finally, the unloading plan and the delivery route are defined from the loading plan and the last random-keys. Similarly as before, it is created iteratively in $n$ steps, where each of one is dependent on the previous one. For each $k \in \{1, 2, ..., n\}$ we map the $\min(n-k + 1, L + 1)$ items that may be delivered without violating the partial LIFO constraints to $\min(n-k + 1, L + 1)$ random-keys. Then, we sort the items in non-decreasing order according to the values of the mapped random-keys. The $\min (n-k + 1, L + 1) - 1$ first items in the order to define the $k$-th container configuration in the delivery area and the last item is then delivered, iteratively making up the delivery route. In Figure~\ref{fig:brkga_decoding_example}, take for example the first step ($k = 1$), where we map items 6, 3 and 2 (only these items may be relocated because in this case, the reloading depth is 2) to the random keys 0.68, 0.94 and 0.95, respectively. After sorting these random-keys, item 2 (highlighted in gray), which is mapped to the greatest random-key (0.95 in this example) is delivered. The other items are stored in the container following the order of their random-keys. This process is repeated until the whole unloading plan has been completed.

\subsection{Initial population}
\label{sec:brkga_initial_population}

Our BRKGA maintains a population $\mathcal{P}$ of $N$ individuals throughout the evolutionary process. To make the initial population, we create $N - 1$ random individuals, where each random-key of each individual is generated independently at random in the real interval $[0, 1]$. Furthermore, to introduce an orientation (hopefully a good one) to the search procedure of the BRKGA, we create and insert into the initial population an individual that represents the initial solution computed to initialize the mathematical models, i.e., the optimal DTSPPL solution in which no rearrangements of items are performed. This last individual is created following the reverse process of the decoding procedure previously described.

\subsection{Biased crossover}
\label{sec:brkga_crossover}

In order to combine the genetic information from the parents and generate new offsprings, a BRKGA uses a biased crossover operator. This crossover always involves two parents, where one is randomly selected from the elite group $\mathcal{P}_e$, and the other is randomly chosen from the non-elite group $\mathcal{P}_{\bar{e}}$. The groups $\mathcal{P}_e$ and $\mathcal{P}_{\bar{e}}$ are formed at each generation of the algorithm after all individuals of population $\mathcal{P}$ have been decoded. Group $\mathcal{P}_e$ is formed by the individuals with greater fitness, while $\mathcal{P}_{\bar{e}}$ is formed by the other individuals, i.e., $\mathcal{P}_{\bar{e}} = \mathcal{P} \setminus \mathcal{P}_{e}$. Moreover, the biased crossover operator has a parameter $\rho_{e}$ that defines the probability of each random-key of the elite parent to be inherited by the offspring individual. More precisely, from an elite parent $a$ and a non-elite parent $b$, we can generate an offspring $c$ according to the biased crossover as follows:\\[-8mm]

\begin{equation}
c_{i} \gets
\begin{cases}
\;a_{i} \quad \quad \text{if} \; \; random\,(0,\,1)\;\leq\;\rho_{e}\\
\;b_{i} \quad \quad \text{otherwise}
\end{cases}
\qquad \qquad \forall i \in \{1, 2, ..., M\} \nonumber
\end{equation}

\noindent where $M$ is the number of random-keys of the individuals and $a_{i}$, $b_{i}$ and $c_{i}$ are, respectively, the $i$-th random-key of individuals $a$, $b$ and $c$.\\[-8mm]

\subsection{Mutant individuals}
\label{sec:brkga_mutants}

Unlike most GAs, BRKGAs do not contain mutation operators. Instead, to maintain population diversity, they use mutant individuals, which are merely new individuals generated by choosing for each random-key a real number between 0 and 1.

\subsection{Next generation}
\label{sec:brkga_next_generation}

In the BRKGA scheme, from any generation $k$, a new population is formed based on the current population $\mathcal{P}$. First, all elite individuals  of generation $k$ in $\mathcal{P}_{e}$ are copied into the new population (generation $k+1$) without any modification. Next, some mutant individuals are added to the new population to maintain high population diversity. Finally, to complete the new population, new individuals are added by using the biased crossover operator.

\subsection{Overall BRKGA}
\label{sec:brkga_overall}

The previous components are organized as described in Algorithm \ref{alg:brkga}. Initially (line~\ref{alg:brkga_init_s}), the best solution found by the algorithm is initialized as an empty solution. Then, at line~\ref{alg:brkga_init_pop}, the initial population is generated. While the stopping criterion is not achieved, the algorithm performs its evolutionary cycle (lines~\ref{alg:brkga_loop_begin} to \ref{alg:brkga_loop_end}). At lines~\ref{alg:brkga_decode_begin} to \ref{alg:brkga_decode_end}, all individuals in the current population are decoded and the best solution found is possibly updated. After selecting the individual elites (line~\ref{alg:brkga_elite_individuals}) and generating the mutant individuals (line~\ref{alg:brkga_mutant_individuals}) as well as the offspring individuals (line~\ref{alg:brkga_offspring_individuals}), the algorithm updates the population of individuals (line~\ref{alg:brkga_update_population}). At the end of the algorithm (line~\ref{alg:brkga_return}), the best solution found is returned.

\begin{algorithm}[!ht]
\DontPrintSemicolon
\SetKwData{Left}{left}
\SetKwData{Up}{up}
\SetKwFunction{FindCompress}{FindCompress}
\SetKwInOut{Input}{input}
\SetKwInOut{Output}{output}
$s^{\texttt{best}} \gets \varnothing$ \label{alg:brkga_init_s} \\
$\mathcal{P} \gets $ initial population with $N$ individuals \label{alg:brkga_init_pop}\\
\Repeat{\upshape time limit is reached} { \label{alg:brkga_loop_begin}
	\ForEach{$p \in \mathcal{P}$} { \label{alg:brkga_decode_begin}
		$s \gets $ individual $p$ decoded\\
		\lIf{\upshape $s$ is better than $s^{\texttt{best}}$} {
			$s^{\texttt{best}} \gets s$ \textbf{  end}
		}
	} \label{alg:brkga_decode_end}
	$\mathcal{P}_{e} \gets $ set of the $N_{e}$ best individuals (elite) from $\mathcal{P}$ \label{alg:brkga_elite_individuals} \\ 
    $\mathcal{P}_m \gets $ set of $N_{m}$ mutant individuals \label{alg:brkga_mutant_individuals} \\
    $\mathcal{P}_o \gets $ set of $N - N_{e} - N_{m}$ offspring individuals \label{alg:brkga_offspring_individuals} \\
    $\mathcal{P} \gets \mathcal{P}_e \,\cup\;\mathcal{P}_{m} \,\cup\;\mathcal{P}_{o}$ \label{alg:brkga_update_population} \\    
} \label{alg:brkga_loop_end}
\Return $s^{\texttt{best}}$ \label{alg:brkga_return}
\caption{Biased Random-Key Genetic Algorithm (BRKGA). }
\label{alg:brkga}
\end{algorithm}%

\section{Computational experiments}
\label{sec:computational-experiments}

In this section, we present the computational experiments performed to study the performance of the proposed solution approaches. As there is no previous work on the DTSPPL, we compare the proposed approaches. Here, we graphically present the results obtained, whereas all numerical results for each instance can be found at 
\href{http://www.goal.ufop.br/dtsppl}{\textcolor{blue}{http://www.goal.ufop.br/dtsppl}}, where is also available our code, as well as all best solutions (tours and loading/unloading plans) found by each solution approach.

Our solution approaches have been coded in C/C++ language. Mathematical models have been solved using Gurobi solver version 9.0.1. The proposed BRKGA has been coded from the framework developed by \cite{toso2015c}. All experiments have been sequentially (nonparallel)  performed on an Intel(R) Xeon(R) E5-2660 (2.20GHz), running under CentOS Linux 7 (Core). 

\subsection{Benchmarking instances}

To assess the quality of the proposed solution methods, we have defined a comprehensive set of 1080 DTSPPL instances. The instances are divided into  216 different types in order to analyze the solution methods on different instance characteristics. Each type is described by the number of customers $n$, the reloading depth $L$, and the cost of each rearrangement $h$. The number of customers $n$ varies in \{6, 8, 10, 12, 15, 20\}, while the reloading depth $L$ and the cost $h$ vary in \{1, 2, 3, 4, 5, $n$\} and \{0, 1, 2, 5, 10, 20\}, respectively. Note that for $h = 0$, we allow rearrangements according to the reloading depth $L$ without any reloading cost. In turn, for $L = n$, we allow any rearrangement of the items, regardless of the number of items introduced in the container. Although this last configuration seems very unpractical in real-world applications, we have considered it in our computational experiments so as to have an interesting comparison term. For each type of instances, we have used the areas R05, R06, R07, R08, and R09 defined by \cite{petersen2009double} for the DTSPMS. Each of these areas consists of two sets, where one defines the pickup locations (pickup region), and the other defines the delivery locations (delivery region). The locations of each region have been generated randomly in a $100 \times 100$ square. The distance between any two points of each region is the Euclidean distance rounded to the nearest integer, following the conventions from the TSPLIB. The first point of each region, which is fixed in coordinates $(50, 50)$, corresponds to the depot, while the next $n$ points define the $n$ customer locations. 

\subsection{Parameter settings}

Our mathematical models ILP1 and ILP2 have been solved by using Gurobi solver with all its default settings. The exceptions are the runtime, which has been limited to one hour, and the optimization process, which has been limited to use only a single processor core. Regarding the parameters of the BRKGA, we have used as stopping criterion the same execution time that both mathematical models have been limited, i.e., one hour. For the other BRKGA parameters, we have used the automatic configuration method I/F-Race \citep{birattari2010f} to find the most suitable configuration. We have used the implementation of I/F-Race provided by the Irace package \citep{lopez2016irace}, which is implemented in the R language and is based on the iterated racing procedure. In our tuning experiments, we have used all Irace default settings, except for the parameter \textit{maxExperiments}, which has been set to 5000. This parameter defines the stopping criterion of the tuning process. We refer the readers to \cite{lopez2016iraceguide} for a complete user guide of the Irace package.

In Table~\ref{table:brkga_parameters}, we describe the BRKGA parameters as well as the tested values for them. Note that the population size $N$ is given in terms of the size of each individual. Moreover, the elite population size $N_{e}$ and mutant population size $N_{m}$ are granted in terms of $N$.  After a vast experiment that used a sample of 10\% of all 1080 instances, Irace pointed out the parameters highlighted in bold in Table~\ref{table:brkga_parameters} as the best ones.\\[-6mm]

\begin{table*}[!ht]
\footnotesize
\centering
\caption{BRKGA parameters.}
\setlength{\tabcolsep}{0pt}
\begin{tabular*}{\hsize}{@{}@{\extracolsep{\fill}}lll@{}}
\toprule
Parameter & Description & Tested values \\ 
\midrule
$N$ & population size & $1M, 2M, 5M, 10M, 20M, 50M, 100M, \boldsymbol{200M}, 500M$ \\ 
$N_{e}$ & elite population size & $0.05N, \boldsymbol{0.10N}, 0.15N, 0.20N, 0.25N, 0.30N$ \\ 
$N_{m}$ & mutant population size & $0.05N, 0.10N, 0.15N, 0.20N,  \boldsymbol{0.25N}, 0.30N$ \\ 
$\rho_{e}$ & elite allele inheritance probability & $0.55, 0.60, 0.65, \boldsymbol{0.70}, 0.75, 0.80, 0.85, 0.90$ \\ 
\bottomrule
\multicolumn{3}{l}{\footnotesize $M$ is the number of random-keys of each individual.}
\end{tabular*}
\label{table:brkga_parameters}
\end{table*}

\subsection{ILP1 $vs.$ ILP2}

\begin{figure}[!ht]
    \centering
    \includegraphics[width=0.90\textwidth]{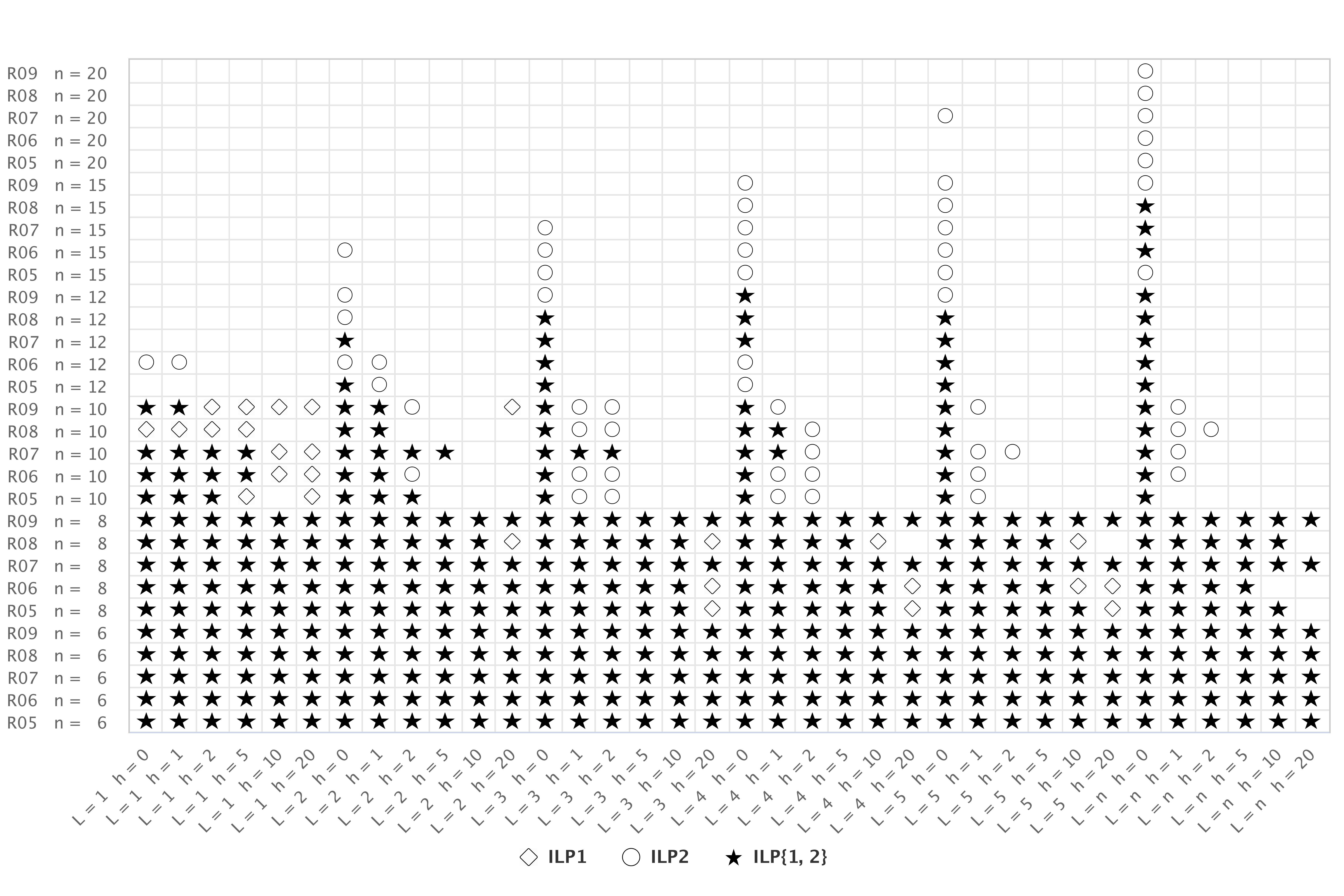}
    \caption{Instances solved to proven optimality.}
    \label{fig:marker_optimal_solutions}
\end{figure}

Our first analysis of results contrasts the mathematical models ILP1 and ILP2. We begin by comparing the performance of each model regarding the optimal solutions found. The results that we obtained are shown in Figure~\ref{fig:marker_optimal_solutions}, where each cell represents a single test instance (horizontal axis informs $L$ and $h$, and vertical axis informs \textit{area} and $n$). We indicate with markers which of the instances have been solved to proven optimality by each model. Empty cells indicate for those instances that no model has been able to prove optimality within one hour of processing time, while markers $\diamond$ and $\circ$ evince those solved by models ILP1 and ILP2, respectively. We point out with $\small\mbox{\ding{72}}$ when both models have been solved to proven optimality.

It can be noticed from Figure~\ref{fig:marker_optimal_solutions} that the proposed mathematical models have been able to solve all instances with 6 customers, and almost all instances with 8 customers. Both models have presented difficulties in solving instances with 10 or more customers, especially with the increase of the reloading depth $(L)$ and the cost of each rearrangement $(h)$. However, we can observe that for instances when there are no rearrangement costs, i.e., $h = 0$, both models perform better with the increase of the reloading depth. Indeed, note that the optimal solutions for instances with $h = 0$ tend to approximate the solution formed by the pickup and delivery optimal tours as the reloading depth increases. This trend results in an easier combinatorial problem, which has been verified in the performance of our models.

\begin{figure}[!ht]
    \centering
    \includegraphics[width=0.90\textwidth]{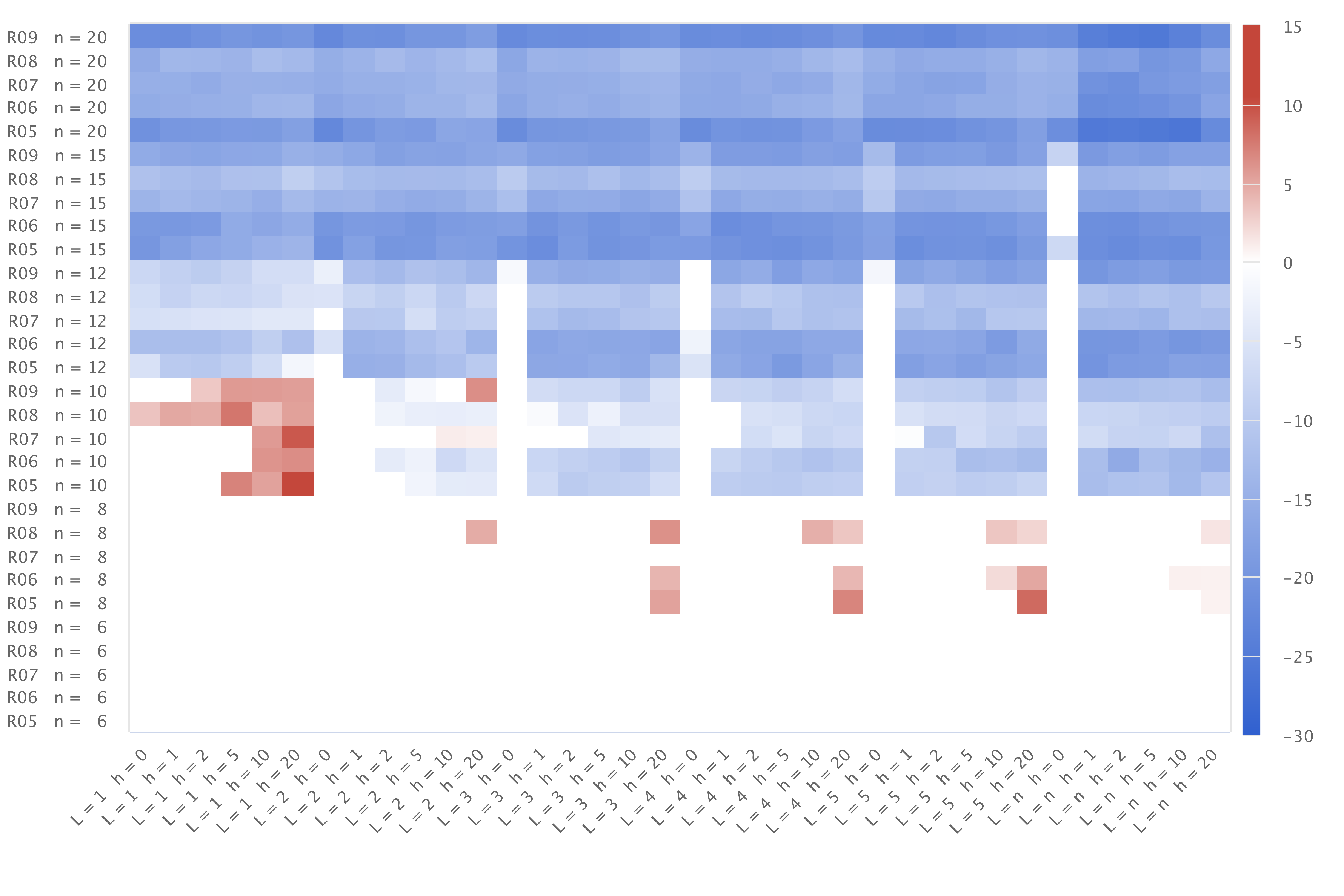}
    \caption{Percentage variation between the lower bounds \texttt{ILP1\_LB} and \texttt{ILP2\_LB}.}
    \label{fig:heatmap_comp_lb}
\end{figure}

To better investigate the behavior of the mathematical models ILP1 and ILP2, we measure the percentage variation of their lower bounds (\texttt{ILP1\_LB} and \texttt{ILP2\_LB}, respectively) reached at the end of computation by calculating \texttt{$($ILP1_LB - ILP2_LB$)$ / max$($ILP1\_LB, ILP2\_LB$)$ $\times$ 100}\%. The percentage variations obtained between the lower bounds are graphically shown in Figure \ref{fig:heatmap_comp_lb}, where we have used a heatmap visualization to emphasize larger variations. Note that positive variation values (highlighted in shades of red) indicate that model ILP1 has reached higher lower bounds than those reached by the model ILP2. In contrast, negative variation values (highlighted in shades of blue) indicate the opposite behavior. Besides, the higher the absolute value (more intense color), the higher the difference between the lower bounds.

From the results reported in Figure~\ref{fig:heatmap_comp_lb}, we can observe that model ILP2 is more effective than ILP1 for most of the instances regarding the lower bounds achieved. It must be stressed that the most instances where the model ILP1 has found tighter lower bounds are those in which have been solved to proven optimality only by that model, as shown previously in Figure~\ref{fig:marker_optimal_solutions}. Besides, we can see that the absolute value of the negative variation is higher than the positive variation, thus indicating, in general, a better performance of model ILP2 concerning the lower bounds. Interestingly, we have noted in our experiments that for many larger instances even the linear relaxation of model ILP2 is tighter than the lower bound reached at the end of one hour of processing time of the model ILP1. We recall the readers interested in the detailed numerical results that we made them available at \href{http://www.goal.ufop.br/dtsppl}{\textcolor{blue}{http://www.goal.ufop.br/dtsppl}} along with our code and solutions.

\begin{figure}[!ht]
    \centering
    \includegraphics[width=0.90\textwidth]{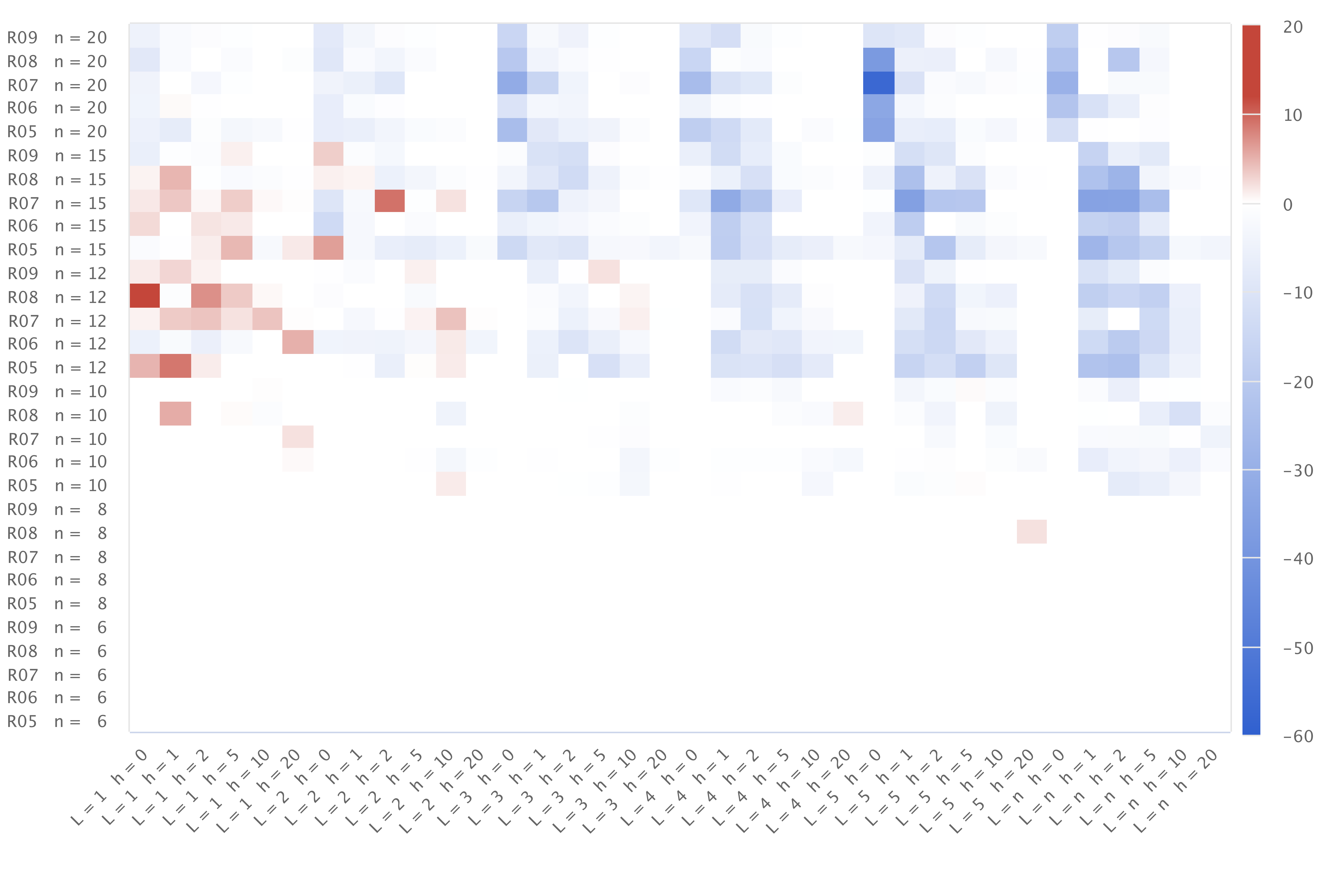}
    \caption{Percentage variation between the upper bounds \texttt{ILP1\_UB} and \texttt{ILP2\_UB}.}
    \label{fig:heatmap_comp_ub}
\end{figure}

We compare now the models ILP1 and ILP2 regarding their upper bounds (\texttt{ILP1\_UB} and \texttt{ILP2\_UB}, respectively) reached at the end of computing. Figure~\ref{fig:heatmap_comp_ub} reports a similar heatmap visualization to the one shown in Figure~\ref{fig:heatmap_comp_lb}. However, each cell now represents the calculated value as \texttt{$($ILP2_UB - ILP1_UB$)$ / min$($ILP1\_UB, ILP2\_UB$)$ $\times$ 100}\%. Note that cells with more intense colors indicate a higher difference between the upper bounds. While positive values (highlighted in shades of red) indicate that model ILP2 has found better solutions than those found by the model ILP1, negative values (highlighted in shades of blue) indicate the opposite behavior. 

Although the models have shown substantial differences concerning their lower bounds, the results reported in Figure~\ref{fig:heatmap_comp_ub} show that both models have obtained solutions with the same and similar quality (white and light blue/red cells), i.e., same and similar upper bounds, for several instances. Disregarding the smaller instances where both models have found optimal solutions, this fact has befallen on many instances that involve large rearrangement costs because the models have not even been able to improve their initial incumbent solution (which we recall is the optimal solution obtained when reloading operations are not allowed). It indicates that in these instances the customers are located in such a way that no rearrangement is attractive or that both models have difficulty finding a way to accomplish them.

For the cases where the models showed significant differences between their upper bounds, we can observe that model ILP1 has performed better for those instances with smaller rearrangement depth. On the other hand, model ILP2 has achieved better solutions for more cases. Furthermore, it has also reached higher absolute differences between the upper bounds found by the model ILP1.

\begin{figure}[!ht]
    \centering
    \includegraphics[width=0.85\textwidth]{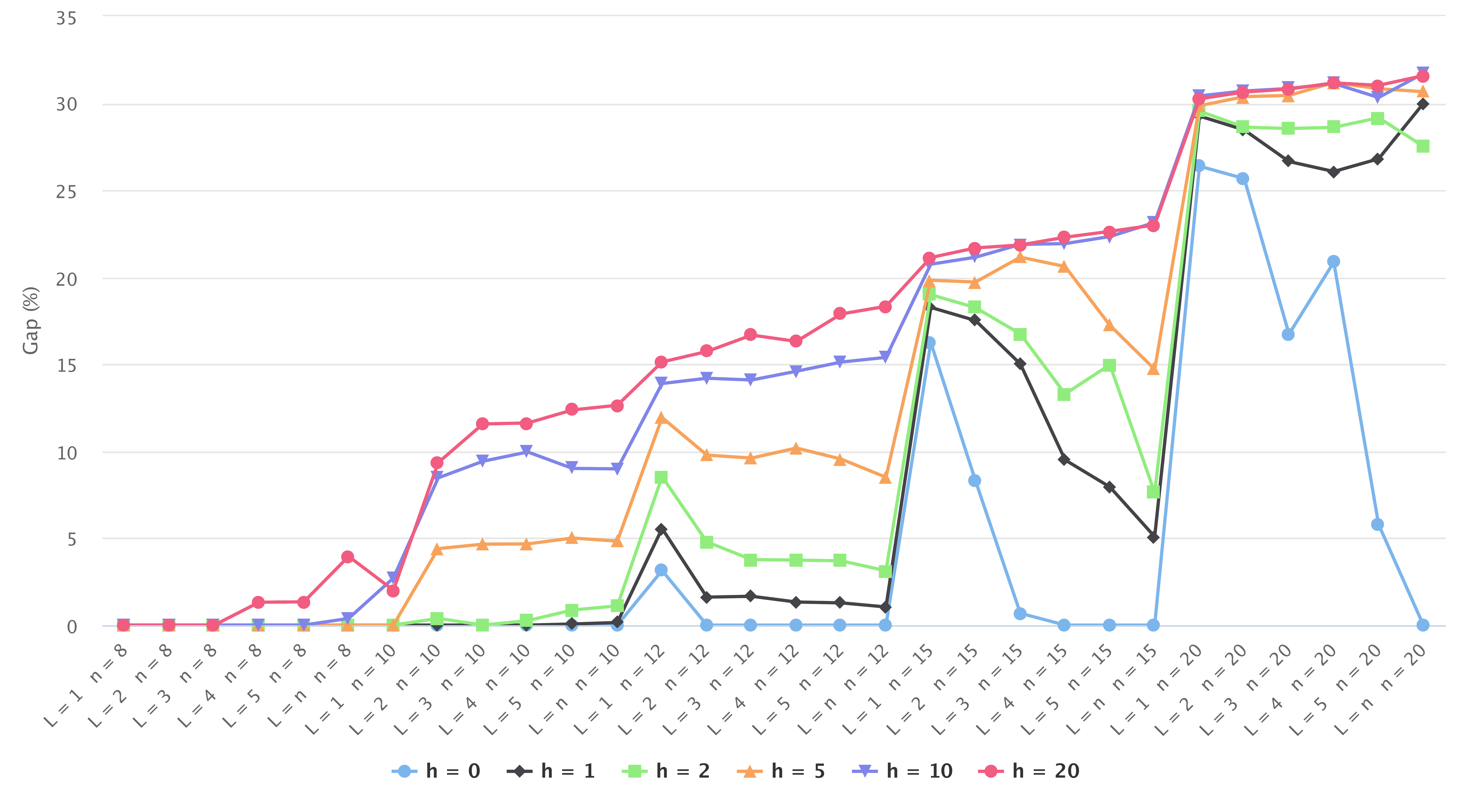}
    \caption{Relative gap between the best lower and upper bounds.}
    \label{fig:gap}
\end{figure}

We conclude the analysis of our models by examining the gap, which is computed as \texttt{$($UB - LB$)$ / UB $\times$ 100}\%, between the best lower and upper bounds (\texttt{LB} and \texttt{UB}, respectively) reached by both models. To be clear, the \texttt{LB} and \texttt{UB} values are calculated as \texttt{max(ILP1_LB, ILP2_LB)} and \texttt{min(ILP1_UB, ILP2_UB)}, respectively. In Figure~\ref{fig:gap}, we compare these values for all types of instances, except for those types that involve 6 customers. For these cases, both models have been able to solve all instances to proven optimality, and, consequently, gap values are zeros. The results reported in the figure consist of the average for all five instances of each type, i.e., average over five instances (R05, R06, ..., R09). better analyze the results, we plot them into six different lines, wherein each of them contains all instances with the same rearrangement cost. 

As can be seen from the gaps reported in Figure~\ref{fig:gap}, the models' performance is strongly influenced by the rearrangement cost. Note that the higher the rearrangement cost, the larger the gap is, with few exceptions. For larger instances, the gap exceeds the value of 30\%, thus indicating the models' difficulty to solve these instances. A closer look reveals, in general, that for rearrangement cost up to 5, that gap decreases as the reloading depth increases. This befalls because rearrangement operations are more attractive in these cases to minimize the total operation costs, reducing the difference between lower and upper bounds.

\subsection{ILPs $vs.$ BRKGA}

We focus now on the computational analysis of the proposed heuristic algorithm. For this purpose, we have run our BRKGA 10 independent times on each instance, and then used the average value (\texttt{BRKGA$_{\texttt{avg}}$}) of the objective function in these runs in our analysis. To assess the quality of our BRKGA, we compare the \texttt{BRKGA$_{\texttt{avg}}$} values with the best upper bound (\texttt{UB}) found by the mathematical models by means of a heatmap schema to emphasize the difference between the quality of the solutions obtained.

In Figure~\ref{fig:heatmap_comp_lb}, a heatmap is reported, where each one of its cells reports the percentage difference between \texttt{BRKGA$_{\texttt{avg}}$} and \texttt{UB} of a specific instance, which is identified as in the heatmaps previously presented. The value of each cell is calculated as \texttt{$($BRKGA$_{\texttt{avg}}$ - UB$)$ / min$($BRKGA$_{\texttt{avg}}$, UB$)$ $\times$ 100}\%. Now, cells with negative values (highlighted in shades of blue) indicate that BRKGA has found, on average, better solutions than those found by mathematical models. In turn, cells with positive values (highlighted in shades of red) indicate a better performance of the models. The higher the absolute value (more intense color) the higher the difference between the quality of the solutions.

\begin{figure}[!ht]
    \centering
    \includegraphics[width=0.90\textwidth]{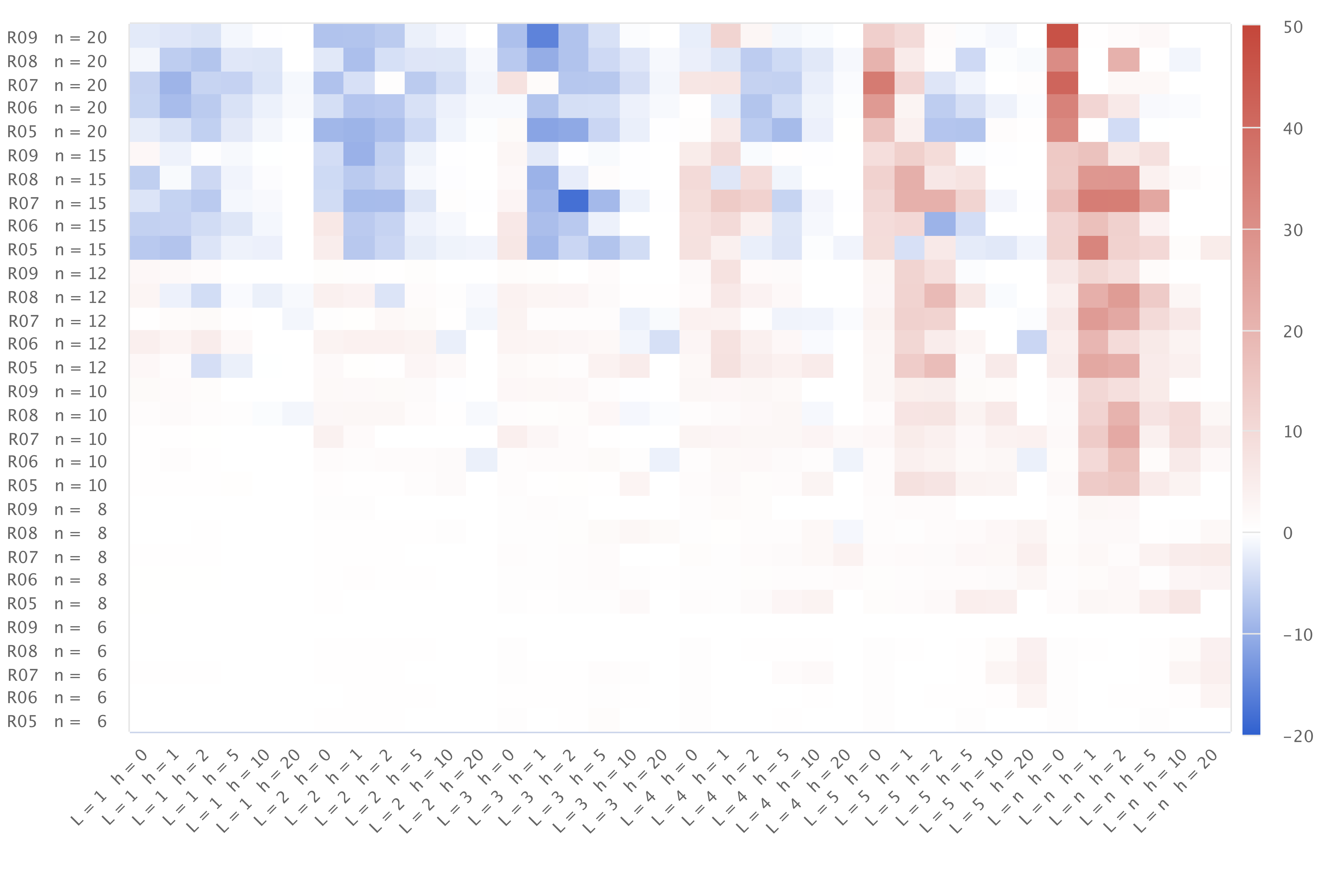}
    \caption{Percentage variation between \texttt{BRKGA}$_{\texttt{avg}}$ and \texttt{UB}.}
    \label{fig:heatmap_ilps_brkga}
\end{figure}

It can be noticed from Figure~\ref{fig:heatmap_ilps_brkga} that our BRKGA has been able to find better solutions than those found by the mathematical models for almost all types of instances that involve reloading depth up to 3. On the other hand, the BRKGA has difficulty dealing with larger reloading depths. For these cases, the models have performed significantly better than our proposed heuristic, although there are some exceptions (see e.g. instances with $L = 5$, $h = \{2, 5, 10\}$, and $n = 20$). Note that the models have performed better for those instances with no limited reloading depth $(L = n)$ and smallest rearrangement costs $(h = \{0, 1, 2\})$. As stated before, for these instances, we have an easier combinatorial problem. Our models, especially the ILP2, take advantage of this behavior, while the BRKGA does not.

\begin{figure*}[!ht]
    \captionsetup[subfigure]{labelformat=empty}
    \centering
    \begin{subfigure}[b]{0.6\textwidth}
        \includegraphics[width=\textwidth]{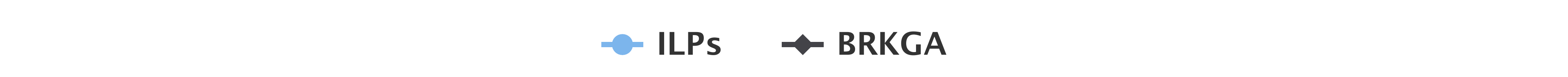}
    \end{subfigure}
    \vspace{0.4cm}
    
    \begin{subfigure}[b]{0.485\textwidth}
        \includegraphics[width=\textwidth]{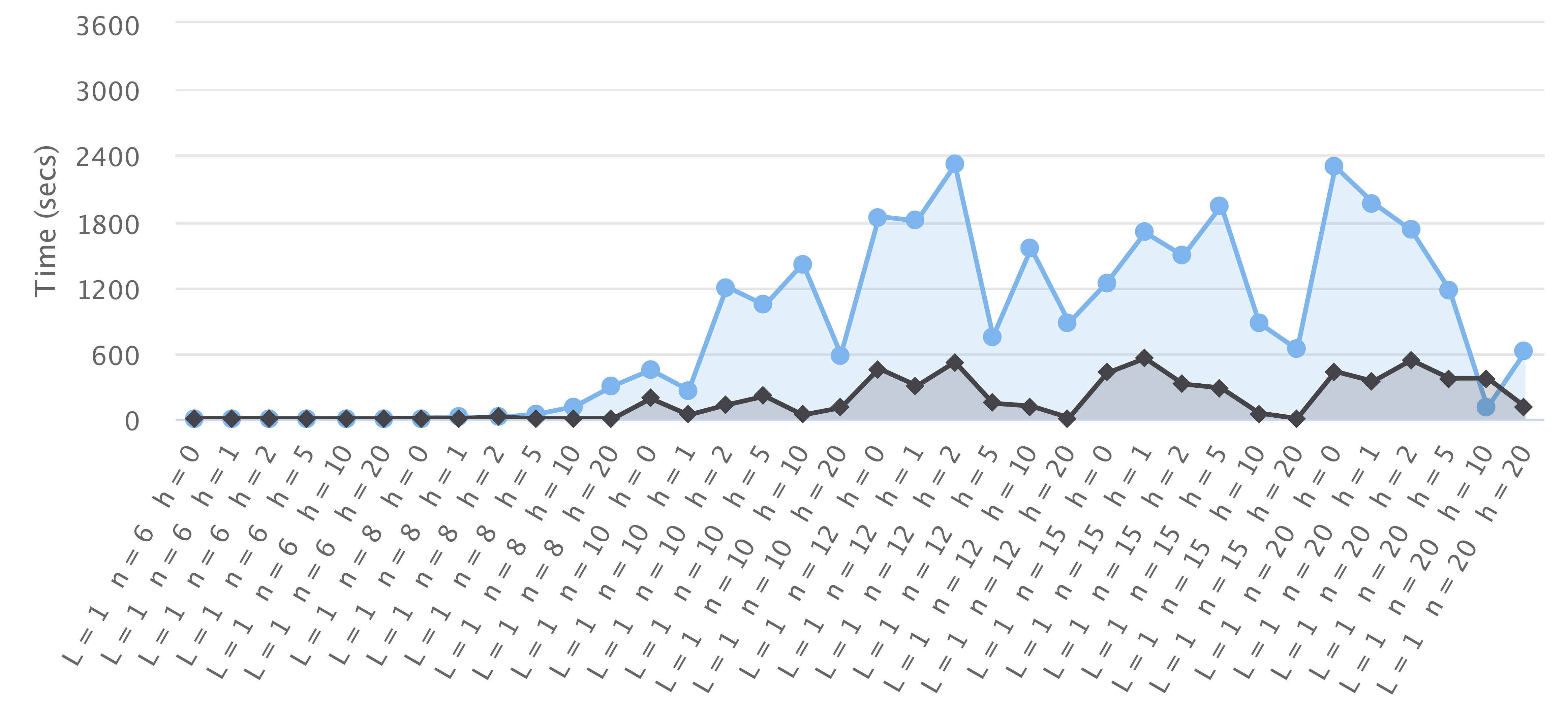}
    \end{subfigure}
    \begin{subfigure}[b]{0.485\textwidth}
        \includegraphics[width=\textwidth]{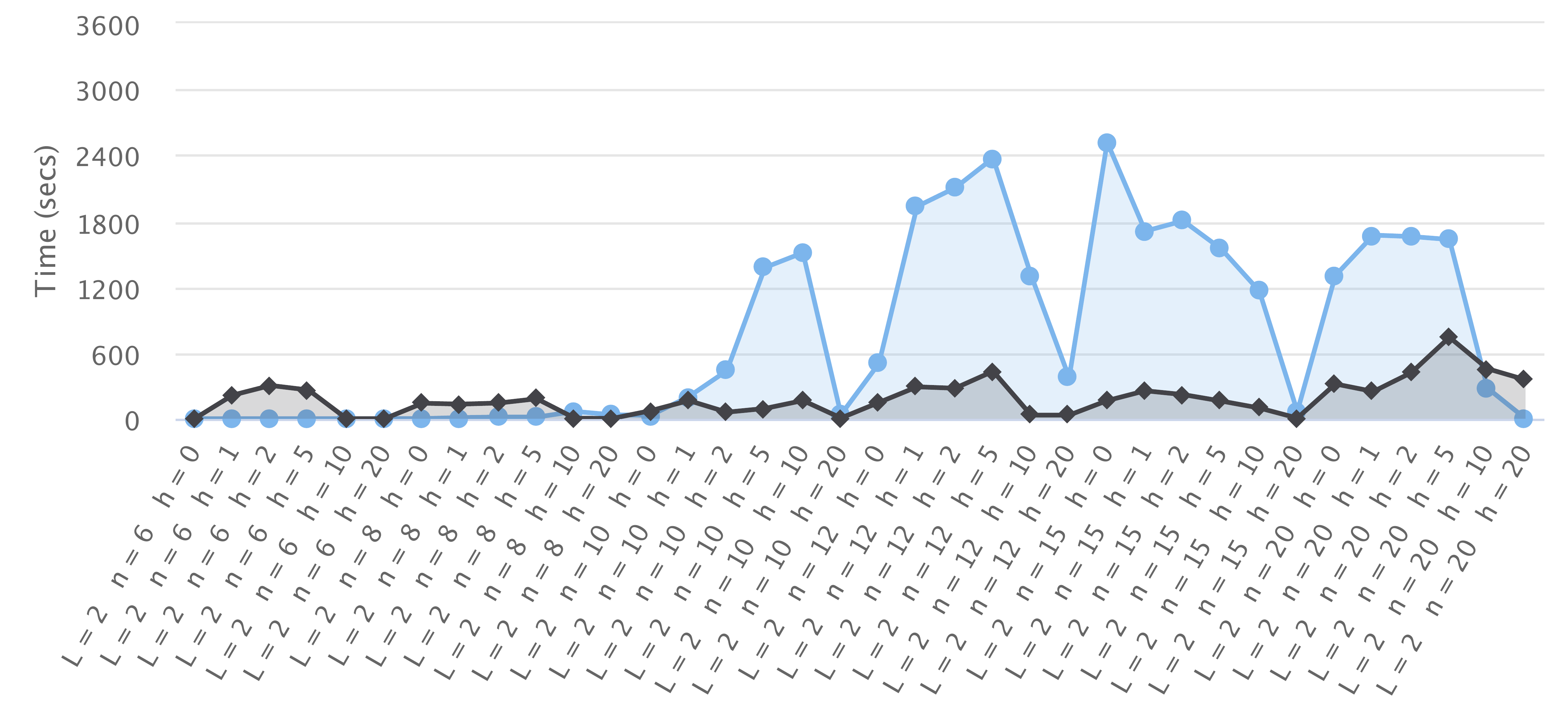}
    \end{subfigure}
    \vspace{0.4cm}
    
    \begin{subfigure}[b]{0.485\textwidth}
        \includegraphics[width=\textwidth]{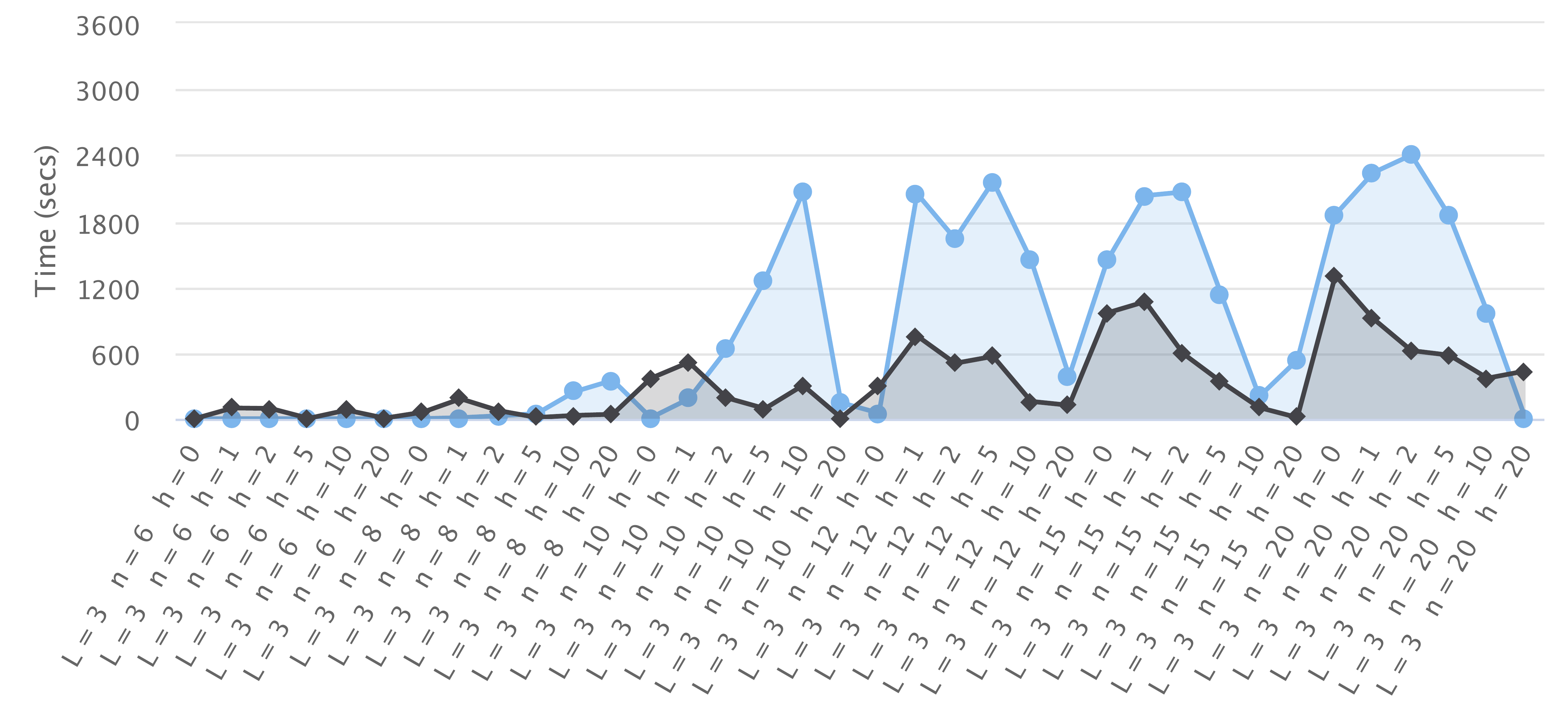}
    \end{subfigure}
    \begin{subfigure}[b]{0.485\textwidth}
        \includegraphics[width=\textwidth]{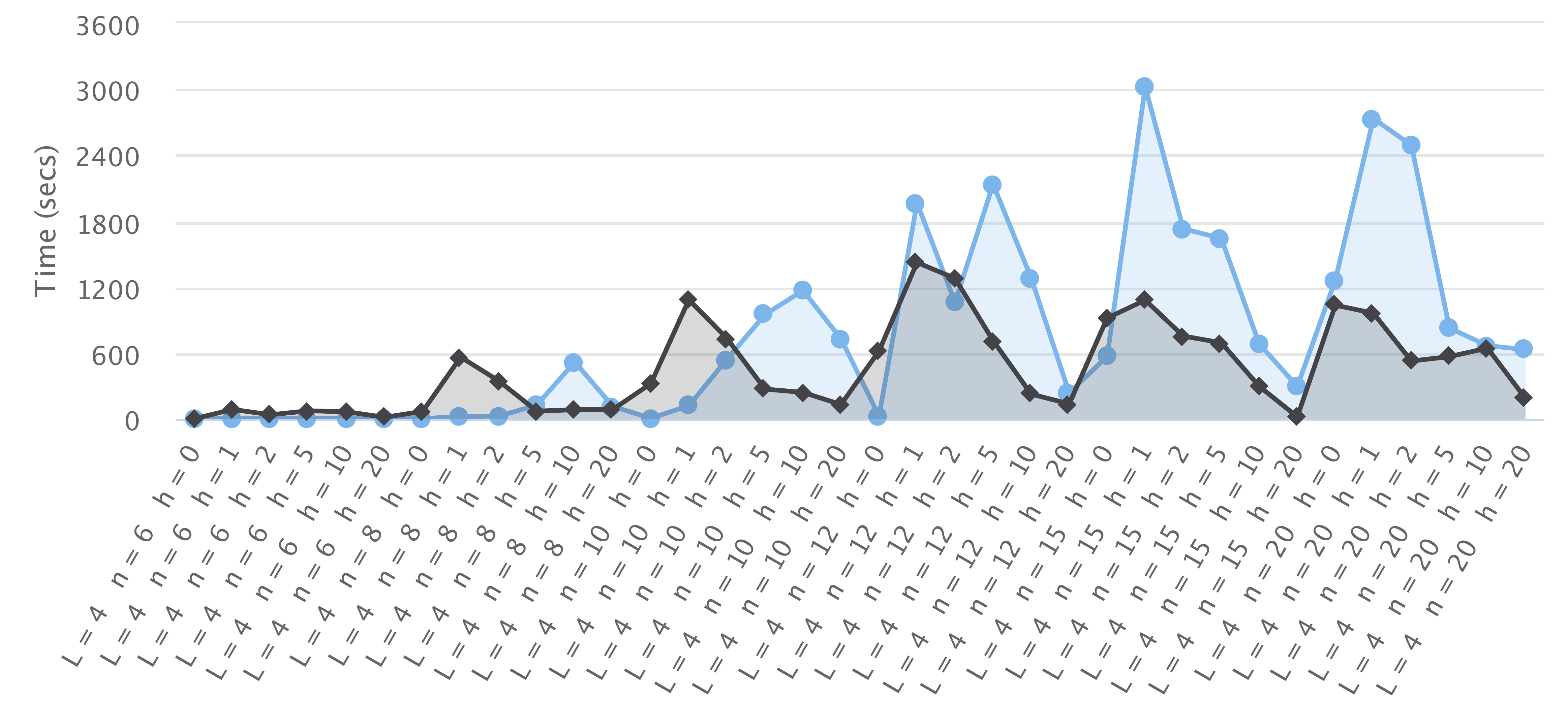}
    \end{subfigure}
    \vspace{0.4cm}
    
    \begin{subfigure}[b]{0.485\textwidth}
        \includegraphics[width=\textwidth]{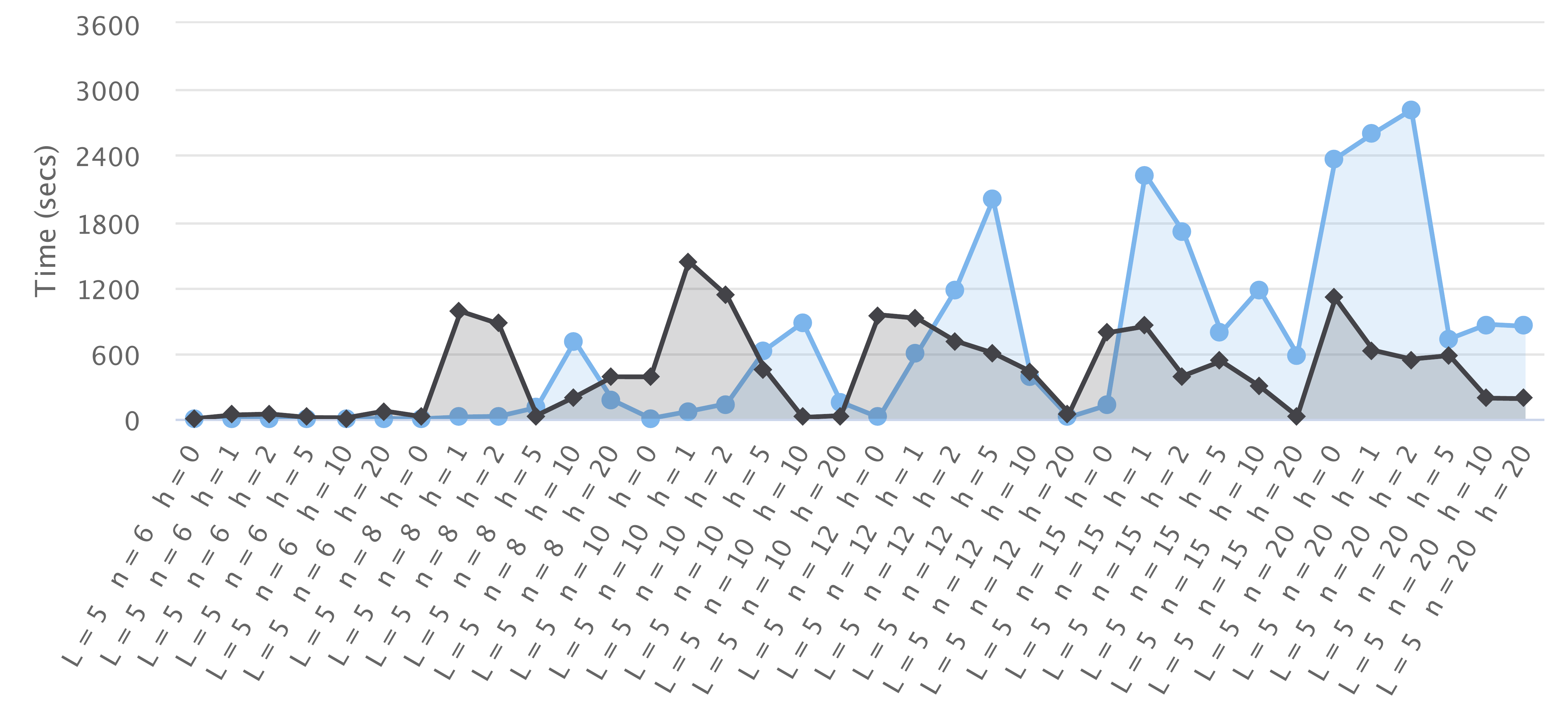}
    \end{subfigure}
    \begin{subfigure}[b]{0.485\textwidth}
        \includegraphics[width=\textwidth]{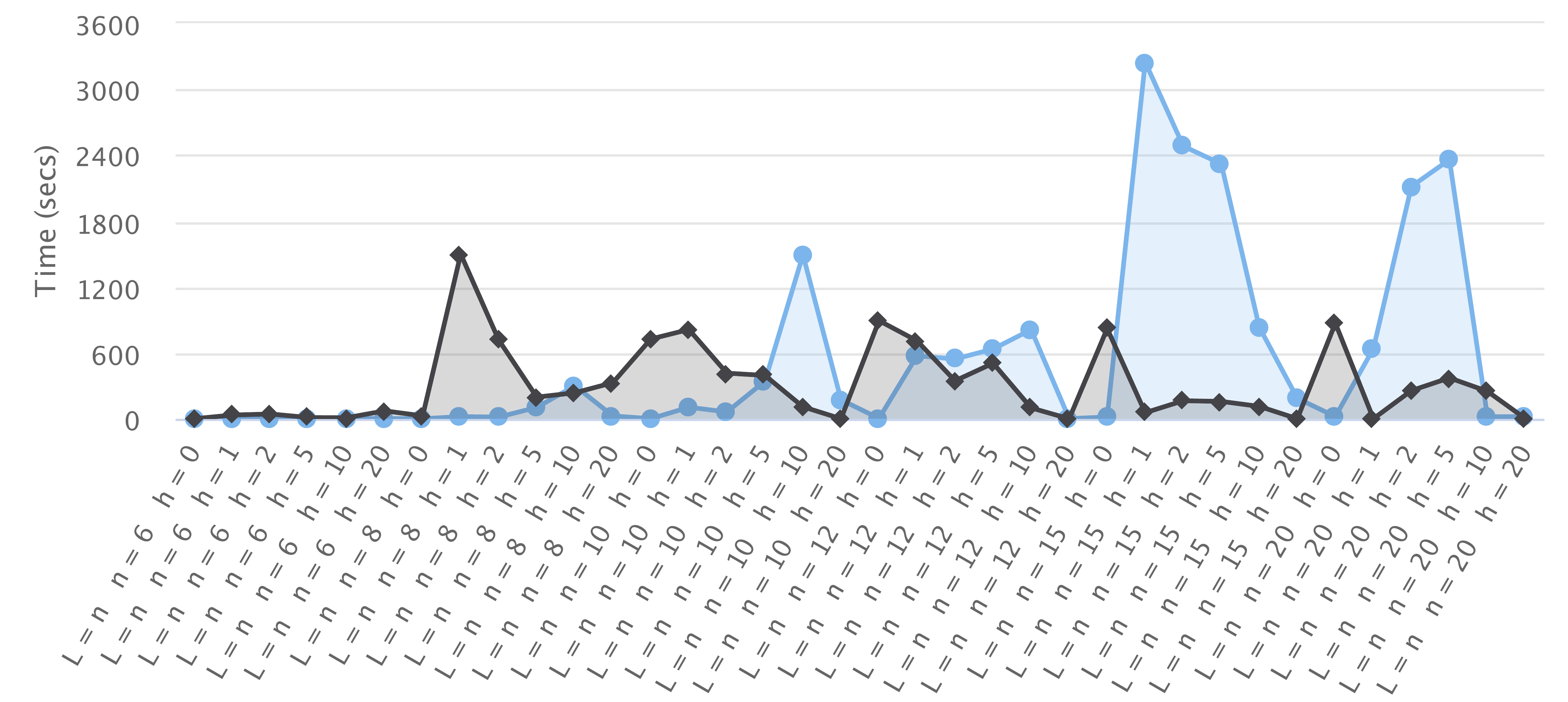}
    \end{subfigure}
    \caption{Computational time to reach \texttt{BRKGA}$_{\texttt{avg}}$ and \texttt{UB} for their solution approaches.} 
        \label{fig:time}
\end{figure*}

Now, we compare the quality of solutions reported in Figure~\ref{fig:heatmap_ilps_brkga} according to the time spent from reaching them. To this end, Figure~\ref{fig:time} shows the average time spent over the five instances of each type. To better analyze, we plot these results into six separate charts, wherein each of them contains all instances with the same reloading depth. 

From Figure~\ref{fig:time}, we can note that BRKGA has had a faster convergence for most of the instances when compared with the models, which has had a positive behavior for instances with reloading depth up to 3 where it has been able to find several better solutions. In general, we can affirm that all our solution approaches have faster convergence behavior as the rearrangement cost increases, which was expected as few or no rearrangements would be interesting for these cases.

\begin{figure}[!ht]
    \centering
    \includegraphics[width=0.85\textwidth]{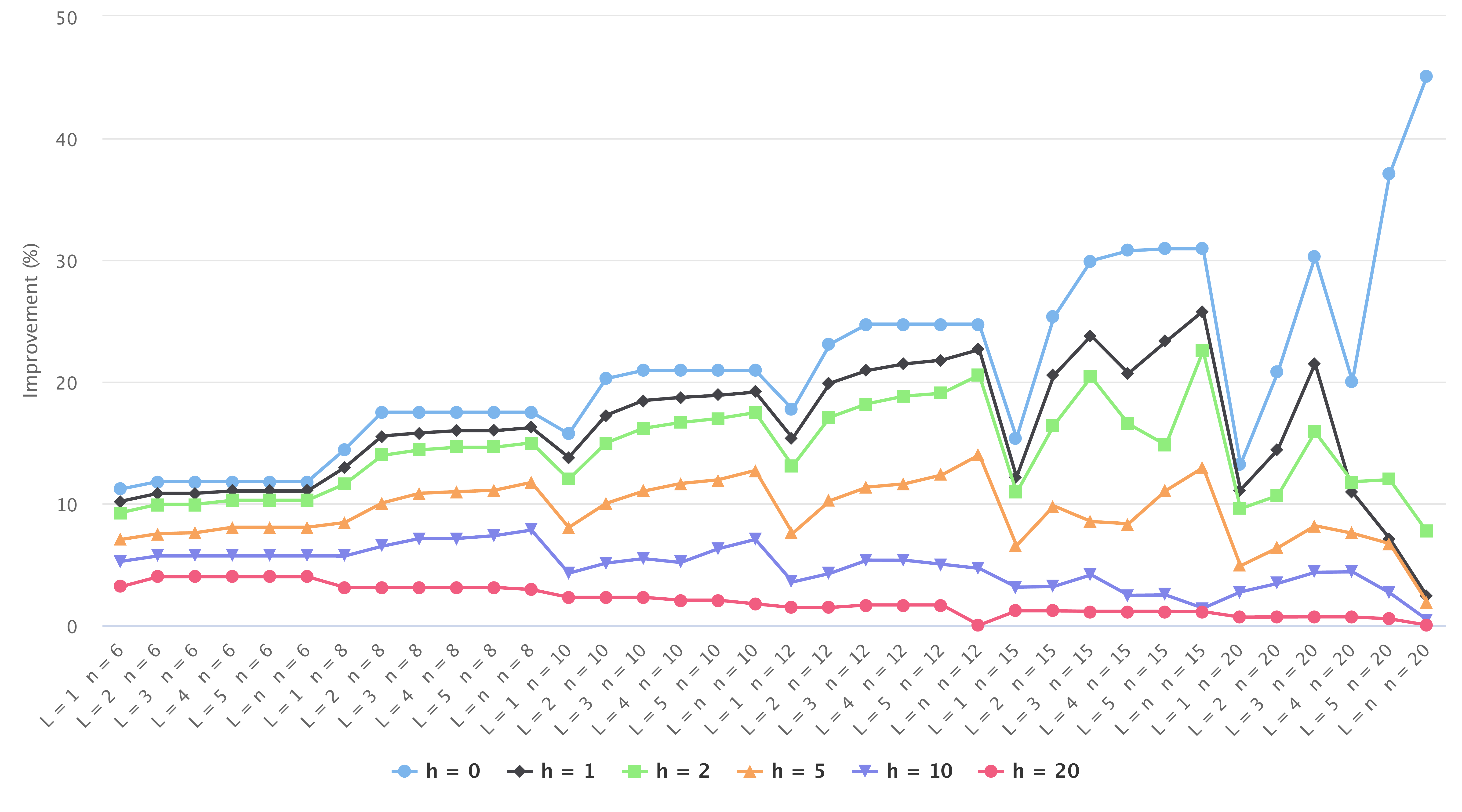}
    \caption{Percentage improvement with partial LIFO loading.}
    \label{fig:ub0_vs_bks}
\end{figure}

Finally, we analyze the percentage improvement of the best solution achieved concerning the optimal solution that does not perform any rearrangement, i.e., when the classic LIFO policy is met. In Figure~\ref{fig:ub0_vs_bks}, we show for each type of instance (a combination of $L$, $n$, and $h$) the percentage improvement in terms of costs obtained by allowing rearrangement items within a reloading depth $L$ and paying a cost $h$ for each item rearranged. Note that all types of instances with the same rearrangement cost are described together in a single line (same color). For example, note that for the type of instance with $L = 2$, $n = 8$, and $h = 5$, we have obtained around 10\% of improvement concerning the optimal solution without rearrangements. As expected, the lower the cost of rearrangement, the higher the percentage of improvement is. Besides, a closer look at the improvement rates shows that when rearrangements are less costly, a deeper reloading limit also implies more significant gains since a higher number of items may be rearranged. Note also that, even when rearrangements are more expensive, some improvements have been reached, especially for smaller instances where our approaches have been able to find solutions with better-proven quality.

\section{Conclusions and open perspectives}
\label{sec:conclusions}

We have approached the Double Traveling Salesman Problem with Partial Last-In-First-Out Loading Constraints (DTSPPL), a pickup-and-delivery single-vehicle routing problem. In this problem, the vehicle has its loading compartment as a single stack, and all pickup and delivery operations must obey a version of the LIFO policy that may be violated within a given reloading depth. We have presented two Integer Linear Programming (ILP) formulations, and we have developed a heuristic algorithm based on the Biased Random-Key Genetic Algorithm (BRKGA) metaheuristic.

The performance of the ILP formulations and the BRKGA has been studied on a comprehensive set of instances built from the DTSPMS benchmark instances. Both ILP formulations have been able to solve to proven optimality only the smaller instances within one hour of processing time. One of them has had tighter lower bounds for almost all instances,  although both formulations have shown similar performance concerning their upper bounds found at the end of the computation. The BRKGA found good quality solutions for all instances, requiring on average short computing times.

There are many possibilities for extending this work. Maybe the most relevant one would be to model and solve the DTSPMSPL, the version of the DTSPPL where the loading compartment of the vehicle is divided into multiple stacks instead of a single one. Because of the increased computational complexity, we expect metaheuristic techniques to be the best option to find good solutions for moderate size instances. It would also be interesting to study the impact of forcing the LIFO policy to be fully respected but allowing at the same time multiple visits to the customers. In such type of problems, iterative aggregate/disaggregate formulations, as in \citep{BI17}, proved to be quite effective. Another important extension of the DTSPPL/DTSPMSPL would be to consider a fleet of vehicles to serve the customers. A larger number of vehicles would cause an increase in the flexibility of the loading/unloading operations, which could lead to a reduction in operating costs. In order to formulate more general problems, would also be interesting to consider partial LIFO loading in situations where backhaul deliveries are dropped, i.e., situations when deliveries are allow without the need to perform all pickup operations before, as in \citep{cote2009branch, cordeau2010branch}. Finally, as multi-objective formulations provide a more powerful optimization tools for decision making, it would be opportune to formulate pickup-and-delivery problems with partial LIFO as bi-objective problems by minimizing the total routing cost and the number of rearrangement operations. 

\section*{Acknowledgments}
The authors would like to thank the anonymous reviewers for their useful comments, which have greatly enhanced this manuscript. The authors thank Coordena\c{c}\~{a}o de A\-per\-fei\-\c{c}o\-a\-men\-to de Pessoal de N\'{i}vel Superior (CAPES) - Finance code 001. The authors would also like to thank Funda\c{c}\~{a}o de Amparo \`{a} Pesquisa do Estado de Minas Gerais (FAPEMIG, grant CEX-PPM-00676-17), Conselho Nacional de Desenvolvimento Cient\'{i}fico e Tecnol\'{o}gico (CNPq, grant 303266/2019-8), Universidade Federal de Ouro Preto (UFOP), Universidade Federal de Viçosa (UFV), and University of Modena and Reggio Emilia (under grant FAR 2018) for supporting this research.

\bibliographystyle{abbrvnat}
\setcitestyle{authoryear,open={((},close={))}}

\bibliography{main}

\end{document}